\documentclass[11pt, a4paper, logo]{lumia}

\usepackage[numbers,sort&compress]{natbib}
\definecolor{infrecred}{HTML}{D2071E}
\definecolor{abstrabg}{HTML}{FCDDDC}
\colorlet{abstradark}{infrecred}
\definecolor{reportaccent}{HTML}{BB0819}
\colorlet{reporttoc}{reportaccent}
\definecolor{cayenne}{HTML}{941100}
\hypersetup{allcolors=cayenne}
\definecolor{darkblue}{rgb}{0, 0, 0.5}
\definecolor{lightblue}{HTML}{D0CFEA}

\usepackage{xspace}

\theoremstyle{plain}

\newtheorem*{proposition*}{Proposition}

\theoremstyle{definition}

\theoremstyle{definition}

\def\eqref#1{equation~\ref{#1}}

\usepackage{float}
\usepackage{multirow}
\usepackage{array}
\usepackage{subcaption}
\usepackage{xurl}
\usepackage{nicefrac}
\usepackage{makecell}
\usepackage{wrapfig}
\usepackage[capitalise,noabbrev]{cleveref}
\usepackage[ruled,vlined,linesnumbered]{algorithm2e}
\crefname{algocf}{Algorithm}{Algorithms}
\Crefname{algocf}{Algorithm}{Algorithms}
\crefname{appendix}{Appendix}{Appendices}
\Crefname{appendix}{Appendix}{Appendices}

\newcommand{\reareport}{}  

\newcommand{\rea}{\textsc{REA}}
\newcommand{\reaM}{\textsc{REA}-M}
\newcommand{\reaA}{\textsc{REA}-A}
\newcommand{\reaV}{\textsc{REA}-V}
\newcommand{\sysz}{System~0}
\newcommand{\syso}{System~1}
\newcommand{\syst}{System~2}
\newcommand{\aspire}{\textsc{ASPIRE}}
\newcommand{\proc}[1]{\textup{\textsc{#1}}}

\setheadertext{REA}

\title{%
  REA: A Routed Embodied Agent Runtime for Asynchronous Brain-Cerebellum Control
}

\setheadertitle{REA: A Routed Embodied Agent Runtime for Asynchronous Brain-Cerebellum Control}

\author{\parbox{\textwidth}{\centering
    \parbox{0.8\textwidth}{\centering
    Pengfei~Zhou$^{1,2}$ \quad
    Yixing~Ma$^{2,3}$ \quad
    Zhengfeiyang~Zhang$^{2,4}$ \quad
    Haoyuan~Chen$^{2,5}$ \quad 
    Zhenglin~Wan$^{1}$ \quad
    Hexin~Wang$^{2}$$^\ddagger$ \quad
    Wangbo~Zhao$^{6}$$^\ddagger$ \quad
    Yang~You$^{1}$$^\ddagger$}\par\vspace{0.35em}
    {\Affilfont
     $^{1}$National University of Singapore, HPC-AI Lab \hspace{4pt}%
     $^{2}$InfRec, Cardinal AI Lab \hspace{4pt}%
     $^{3}$University of California, Berkeley \hspace{4pt}%
     $^{4}$The University of Tokyo\par
     $^{5}$Shenzhen Institutes of Advanced Technology, Chinese Academy of Sciences \quad
     $^{6}$Hong Kong University of Science and Technology\par}}}

\correspondingemail{{}\makebox[\linewidth][l]{\makebox[\dimexpr\linewidth+24pt\relax][c]{\emailicon~\href{mailto:wanghexin@infrec.tech}{wanghexin@infrec.tech}; \emailicon~\href{mailto:wangbo.zhao96@gmail.com}{wangbo.zhao96@gmail.com}; \emailicon~\href{mailto:yangyou@nus.edu.sg}{yangyou@nus.edu.sg} \quad $^\ddagger$ Corresponding Author}}\\[2pt]\makebox[\linewidth][l]{\makebox[\dimexpr\linewidth+24pt\relax][c]{Project page: \href{https://mikeeeeeeeeeeeeee.github.io/REA/}{\texttt{https://mikeeeeeeeeeeeeee.github.io/REA/}}}}}

\begin{document}

\begin{abstract}
Foundation models are becoming remarkably reliable at high-level reasoning, yet they remain too slow and costly to serve as fine-grained controllers. Conversely, fast policies like VLAs and whole-body controllers offer more efficient reflexes, but struggle to generalize. 
While pairing a deliberative brain (foundation models) with a reactive cerebellum (fast policies) seems natural for embodied systems, two obstacles remain: methods are built and scored on incompatible stacks, and their asynchronous coordination is difficult. 
For example, an agent that calls the VLA in the agentic loop can stall the fast execution, while a VLA that calls the agent on errors tends to re-trigger fallbacks repeatedly or completely miss silent failures. 
To address this, we propose REA (Routed Embodied Agent), a unified runtime where a deliberative System 2, an embodied router System 1 and reactive System 0 policies cooperate under a single golden rule: no running policy waits on the brain. 
Instead of querying the brain at each failure, REA employs a \textit{route memory} that stores each policy’s measured success ahead of time, a \textit{boundary-triggered judge gate} that consults the brain on policy execution, and a shared-budget \textit{escalation ladder} for bounded recovery. 
Evaluated on 3,000 test episodes across 60 LIBERO-Pro tasks, REA reaches 95.9\% success and beats the best single-agent baseline in its own policy pool (94.7\%) while reducing control steps by 25\% and control time by 17\%. 
In a separate 100-episode RoboTwin~2.0 study, REA achieves 84\% success with VLA policies and 86\% with VLA+code, compared with 80\% for the strongest single VLA baseline MEDex. These results demonstrate learned and programmatic policies operating within the same runtime. REA also transfers to BEHAVIOR-1K and a real Unitree G1 humanoid, achieving the highest observed success among the compared methods.
\end{abstract}

\teaserfigure{%
  \par\noindent
  \begin{minipage}{\linewidth}
    \setlength{\parskip}{0pt}
    \centering
    \includegraphics[width=\linewidth]
      {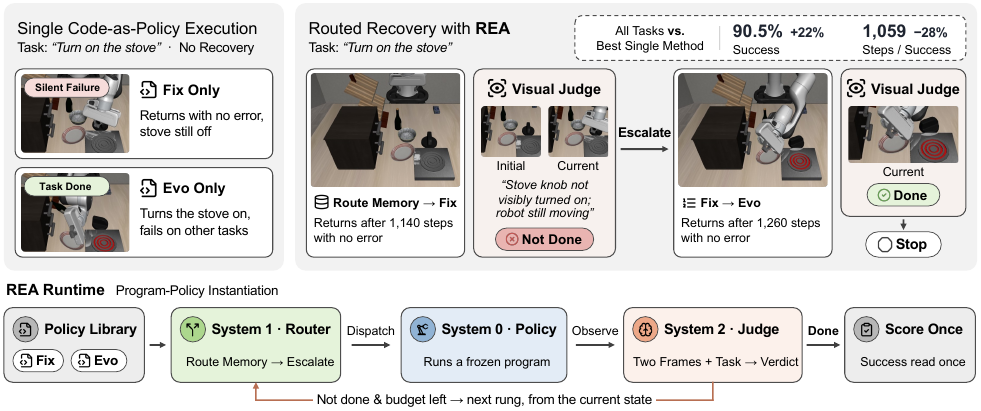}\par

    \captionsetup{font=small,skip=10pt}
    \captionof{figure}{%
      \textbf{Overview of \rea{}.}
      \sysz{} runs generated code policies;
      \syso{} selects them with a route table (\emph{route memory});
      \syst{} is the agentic \emph{judge gate} that reasons and predicts based on current state. Fix and Evo refer to different code-as-policy strategies.
      \label{fig:system}%
    }
  \end{minipage}\par
}

\maketitle

\clearpage
\begingroup
\hypersetup{linkcolor=reporttoc,linktoc=section}
\setlength{\parskip}{0pt}
\small
\setcounter{tocdepth}{2}
\tableofcontents
\endgroup
\clearpage

\section{Introduction}
\label{sec:intro}

\ifdefined\reareport\else 
\begin{figure}[t]
  \centering
  \includegraphics[width=\textwidth]{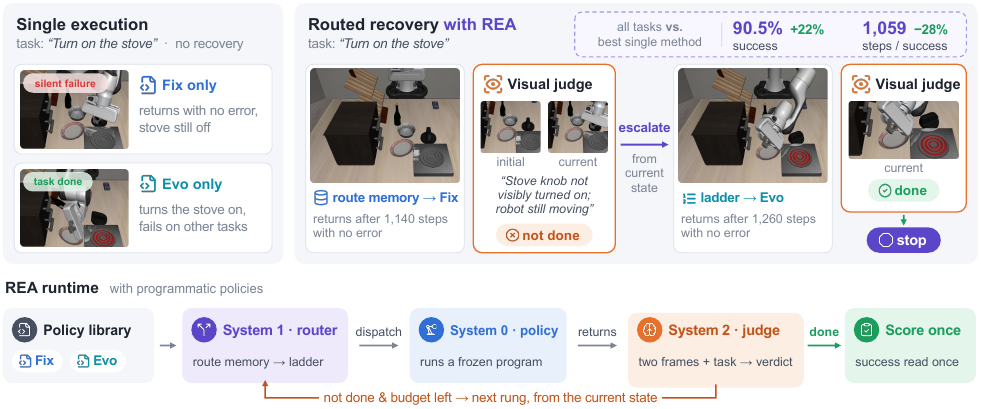}
  \caption{\textbf{Overview of \rea{} with programmatic policies.}  Top left: a single policy may end with the task unfinished (Fix), and a stronger policy is expensive to synthesise (Evo).  Top right: the \syst{} visual judge compares the episode-initial and current frames with the instruction, answers \emph{not done}, and the router dispatches the next policy from the current state.  Bottom: \sysz{} runs frozen programs generated offline; \syso{} selects them with a route table (\emph{route memory}) and a first/other/first \emph{escalation ladder}; \syst{} is the visual \emph{judge gate}.  All runs share one budget; success is read once, after stopping, for evaluation only.}
  \label{fig:system}
\end{figure}
\fi

Foundation-model agents have become capable \emph{embodied brains}: a coding agent can write a manipulation program that calls segmentation, grasp planning and inverse kinematics \citep{liang2023codeaspolicies,aspire2026}, and a language or vision--language model can decompose a command \citep{ahn2022saycan,huang2022innermonologue,driess2023palme} or tell whether a task succeeded \citep{driess2023palme,du2023success}.  These abilities are expensive: in our runs \aspire{}'s repair loop~\citep{aspire2026} spent a median of 43\,min per LIBERO task, and a Codex + GPT-6 pipeline consumed 89M input tokens to generate programs for 60 LIBERO-Pro tasks.  Nor can the brain simply act as the controller: in our probes, models emitting one-token or Cartesian IK actions solved 1 of 1{,}987 and 0 of 600 LIBERO-Pro episodes (\Cref{app:onetoken}, \Cref{fig:placement}).  Vision--language--action models (VLAs) \citep{brohan2023rt2,kim2024openvla,octo2024,black2024pi0} and whole-body controllers such as SONIC \citep{luo2025sonic} act at a few to tens of hertz, far faster than an agent deliberates, but VLAs that exceed 90\% on LIBERO collapse under LIBERO-Pro's perturbed test set \citep{zhou2025liberopro}.

Embodied intelligence should therefore separate a deliberative \emph{brain} from a reactive \emph{cerebellum}, as recent dual-system robot models do \citep{bjorck2025gr00t,figure2025helix,figure2026helix02,shi2025hirobot,gemini2025robotics15}. However, two obstacles still stand in the way.  First, \textbf{there is no common runtime for comparing agents and VLAs}: agent- and VLA-based systems are each built on their own simulator, robot, observation format and success criterion, so it is hard for them to test on the same episodes. Second, \textbf{asynchronous coordination is hard} (\Cref{fig:timeline}).  When the agent drives and calls the fast policy, the fast loop waits on every agent decision: an agent advisor that the router waits for leaves the robot idle for 2.6s per LIBERO-Pro episode, 20\% of its control time (\Cref{fig:speed}).  When the fast policy drives and calls the agent on errors, a persistent error re-triggers the brain at every recurrence, while most failures raise no error at all: an exception-triggered router escalated only 6 times in 590 LIBERO-Pro episodes although 166 failed. Re-execution is not always reliable either: if the task has already succeeded, running it again may undo the result.

\begin{figure}[t]
  \centering
  \includegraphics[width=\textwidth]{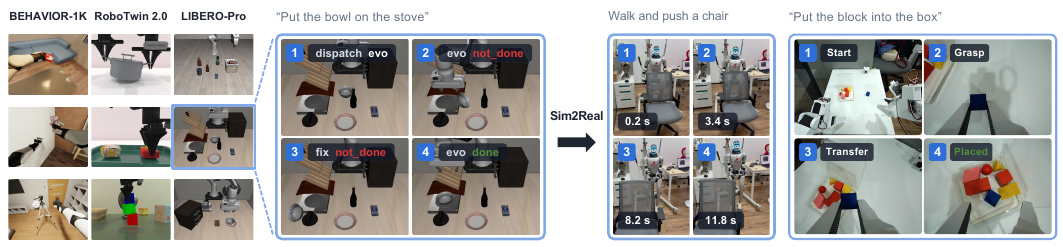}
  \captionsetup{font=small}
  \caption{\textbf{\rea{} across embodiments.} Left: episodes in BEHAVIOR-1K, RoboTwin~2.0 and LIBERO-Pro. Middle: route memory and the judge gate recover from silent failures on a LIBERO-Pro test episode. An Evo--Fix--Evo sequence places the bowl on the stove after the judge rejects its first two attempts; no attempt raised an exception. Right: the same runtime makes a real Unitree G1 walk to and push a chair (8/10 trials).
}
  \label{fig:showcase}
\end{figure}

\begin{figure}[t]
  \centering
  \includegraphics[width=\textwidth]{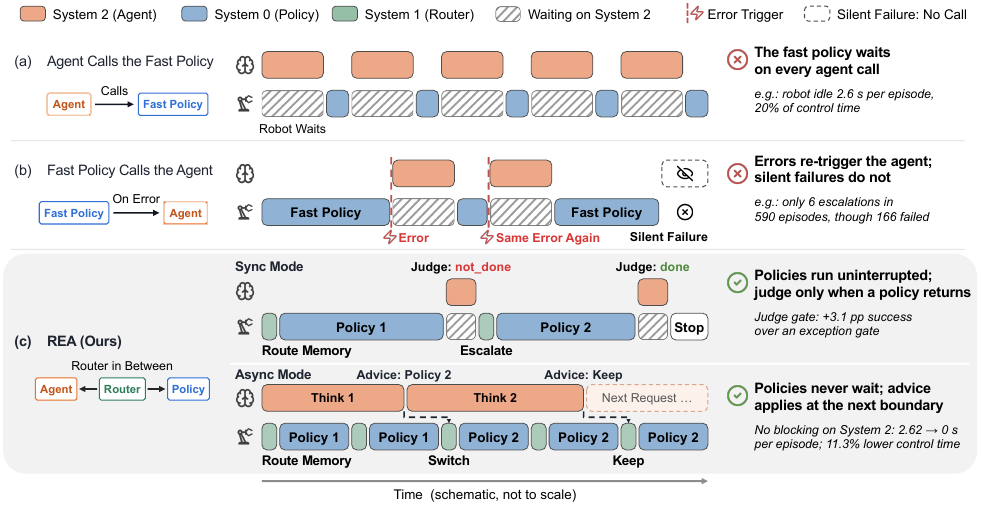}
  \caption{\textbf{Three ways to couple a slow brain with fast policies.} (a) An agent that calls the fast policy puts its latency on every step. (b) A fast policy that calls the agent on errors is interrupted by persistent errors and never by silent failures. (c) \rea{} dispatches from route memory, lets each policy run uninterrupted at native rate, and consults the \syst{} judge only when there is a need.}
  \label{fig:timeline}
\end{figure}

We address the above issues with \rea{}, a \emph{Routed Embodied Agent} runtime (\Cref{fig:system}) designed to unify agent and VLA runtime across simulators and real robots. REA implements a three-level system: \textbf{\syst{}, the brain}, is an agent that generates policies offline or online, and provides advises or judges; \textbf{\syso{}, the embodied router}, decides which policy to run and when to consult the brain without blocking a running policy;  \textbf{\sysz{}, the cerebellum}, holds fast policies (frozen programs, VLA policies or whole-body controllers) that return control to the router at segment or policy boundaries. \rea{} also provides a unified evaluation framework for any policies including code-as-policies and VLAs. Figure~\ref{fig:showcase} shows \rea{} across simulators and on a real robot.

Our central idea is to let the brain prepare policies offline and intervene online only when there is a need, so that no running policy has to wait for it.  Besides the conventional serial execution mode, our \rea{} also offers \emph{asynchronous} mode: policies run at their own rate, with \syst{} running in parallel to provide advice or judge. 
Our proposed method achieves this via three designs. \textbf{To keep the brain off the dispatch path}, \rea{} contains a \textbf{route memory} of each policy's per-task success and cost on development episodes, so choosing a policy can be isolated from the \syst{}'s reasoning process and call \syst{} on dilemma. 
\textbf{To catch silent failures without re-firing on errors}, a \textbf{boundary-triggered judge gate} consults \syst{} concurrently and decides whether a task is finished: if \emph{not done} then escalates, if \emph{done} then stops and protects completed work.  \textbf{To keep recovery bounded}, an \textbf{escalation ladder} dispatches the next policy from the current state under one shared step and time budget.

\paragraph{Results.}
\emph{Routing efficiency.} On 3{,}000 test episodes across all 60 LIBERO-Pro tasks, \rea{} reaches 95.9\% success, exceeding the strongest individual baseline, Agents SDK + GPT-6 (94.7\%; $p{=}0.004$), and coming within 1\% of a hindsight per-task oracle. It is the fastest timed method, with 17\% less control time than Agents SDK + GPT-6 and 25\% less than a blocking advisor. \emph{Failure recovery.} \rea{} improves \aspire{} and zero-shot GPT-6 policies by $+8.6$\% and $+7.0$\%. Judge gating adds $+3.1$\% over exception gating ($p{=}0.022$), at 2.1$\times$ the time per success. \emph{VLA and code policies.} In a separate RoboTwin~2.0 study with 50 clean and 50 randomized episodes, \rea{} achieves 84\% with VLA policies and 86\% with VLA+code, compared with 80\% for MEDex, the strongest single VLA baseline (\Cref{tab:robotwin_vla}). \emph{Transfer.} \rea{} solves 27/50 BEHAVIOR-1K episodes versus the Agents SDK agent's 25/50 (not statistically significant), 97/200 RoboTwin scenes versus Codex CLI's 75/200 in the separate side-by-side study, and 8/10 real Unitree G1 trials.

\paragraph{Contributions.}
(1) \rea{}, a three-level runtime with a common contract for agents, VLAs and programmatic policies, evaluated across three simulators and real robots. (2) An embodied router combining route memory, a boundary-triggered judge gate and a shared-budget escalation ladder to coordinate slow brains and fast policies without blocking execution. (3) Over 30{,}000 aligned test episodes verifying the contributions of routing and judging, plus a separate RoboTwin~2.0 study comparing VLA policies and a mixed VLA--code pool within the same runtime.

\section{Related Work}
\label{sec:related}

\paragraph{Embodied brains and fast policies.}
Large models serve as the deliberative layer of robot systems: SayCan~\citep{ahn2022saycan} grounds LLM plans in
learned affordances, PaLM-E~\citep{driess2023palme} embeds observations into a language model, and Inner
Monologue~\citep{huang2022innermonologue} and SayPlan~\citep{rana2023sayplan} re-plan from feedback.  Code-as-policy
methods let the agent write the controller: Code as Policies~\citep{liang2023codeaspolicies} and
ProgPrompt~\citep{singh2022progprompt} compose perception and motion primitives, Voyager~\citep{wang2023voyager} iterates on executable policy code with environment feedback (Eureka~\citep{ma2023eureka} does so for reward code), and \aspire{}~\citep{aspire2026} scales repair and evolutionary search to LIBERO; more generally, ReAct~\citep{yao2023react} interleaves reasoning with actions and Reflexion~\citep{shinn2023reflexion} adds self-reflection across trials.  At the other end, vision--language--action models such as RT-2~\citep{brohan2023rt2}, OpenVLA~\citep{kim2024openvla}, Octo~\citep{octo2024} and $\pi_0$~\citep{black2024pi0} act far faster than an agent deliberates but generalise within their training distribution~\citep{zhou2025liberopro}, and whole-body controllers such as SONIC~\citep{luo2025sonic} track motions commanded from above.  \rea{} keeps the agent as \syst{}, off the control loop and the dispatch path,
and lets any fast policy serve as \sysz{}.

\paragraph{Dual-system control.}
Following \citet{kahneman2011thinking}, GR00T~N1~\citep{bjorck2025gr00t} and Helix~\citep{figure2025helix} train a slow VLM with a fast action head that run at fixed, different rates, and Helix~02~\citep{figure2026helix02} adds a learned
1\,kHz whole-body ``System~0''; Hi Robot~\citep{shi2025hirobot} and Gemini Robotics~1.5~\citep{gemini2025robotics15} let
a VLM planner command a VLA, and real-time chunking~\citep{black2025rtc} overlaps a VLA's inference with its execution.
Every level there is a learned model of one stack.  \rea{} keeps the level names but not their semantics: \syso{} is a
router that emits no actions, any policy can serve as \sysz{}, and \syst{} is an independently developed agent that
no running policy waits for.

\paragraph{Routing, evaluation and completion judging.}
FrugalGPT~\citep{chen2023frugalgpt} cascades LLM APIs, \citet{lu2023routing} learn reward-guided routers,
mixture-of-experts layers~\citep{shazeer2017outrageously} gate per token, and Agent-as-a-Router~\citep{acrouter2026} routes each coding task using the verified outcomes of similar past tasks; \rea{}'s route memory brings outcome-based routing to embodied policies.  Embodied Agent
Interface~\citep{li2024eai} and SIMPLER~\citep{li2024simpler} each evaluate one family of methods, LLM planners or VLAs,
whereas \rea{} runs brains, routers and policies of different families on identical episodes.  VLMs serve as success
detectors~\citep{du2023success} and reward models~\citep{sontakke2023roboclip}, LLM judges score model
outputs~\citep{zheng2023llmjudge}, and REFLECT~\citep{liu2023reflect} turns detected failures into corrections;
\rea{} proposes no new detector, only where one sits: once per policy return, gating escalation under a shared budget
and never used as the success criterion.

\section{The REA Runtime}
\label{sec:method}

\rea{} answers one question: \emph{how can a slow, general brain and fast, narrow policies cooperate without
the brain ever blocking a running policy?}

\paragraph{REA variants.}
\rea{} denotes the shared runtime. Its three variants in experiments differ in policy selection and recovery, see details in \Cref{alg:rea,app:alg}:
\begin{itemize}[leftmargin=*]
  \item \textbf{\reaM{} (route memory).} Uses route memory to select policies and the exception gate to trigger recovery after an exception or truncation, with no online \syst{} calls.
  \item \textbf{\reaA{} (agent routing).} Calls a \syst{} agent to select the initial policy and choose a recovery policy whenever the exception gate triggers.
  \item \textbf{\reaV{} (visual judging).} Uses route memory to select policies and a visual judge at policy returns to decide whether to stop (\texttt{done}) or advance along the escalation ladder (\texttt{not\_done}).
\end{itemize}

\subsection{Three-Level Contract}
\label{subsec:levels}

An episode is a task $t$ with instruction $\ell_t$, an initial state indexed by seed $s$, a step horizon $H$
and a wall-clock budget $W$.

\paragraph{\sysz{}: policy execution.}
A policy $p$ is any controller that can start from the current state and eventually returns,
$(\Delta,\delta,e,\tau)\leftarrow\proc{Run}(p;\,H-u,\,W-w)$, where $u,w$ are the steps and seconds already
spent, $\Delta,\delta$ those of this run, $e$ an exception or $\bot$ and $\tau$ a truncation flag.  A policy may
also yield at \emph{segment} boundaries and be resumed: a VLA action chunk, a motion interval, or a program
that pauses every 100 control steps with its stack preserved.  Policies run their own closed loop at native
rate.  On LIBERO-Pro they are frozen Python programs written by coding agents; they call SAM~3
\citep{carion2025sam3}, ContactGraspNet~\citep{contact_graspnet}, PyRoki inverse kinematics~\citep{kim2025pyroki} and blocking
motion primitives inside a sandbox that rejects identifiers (such as \texttt{getattr}) that could reach
simulator state. On RoboTwin~2.0, the VLA extension also executes learned action-chunk policies and an online code policy through this interface (\Cref{subsec:vla}).

\paragraph{\syst{}: the brain, offline and online.}
Offline, \proc{Generate}$(t)$ produces candidate policies $\mathcal{P}_t$ (the ``Frozen programs'' box of \Cref{fig:system}).  Online, the brain offers two services: \proc{Judge}$(f_0,f_k,\ell_t)\in\{\texttt{done}, \texttt{not\_done}\}$ compares the episode-initial and current frames with the instruction, and \proc{Advise}$(o)$ recommends a candidate from text inputs (used by \reaA{} at dispatch and by the blocking-advisor baseline). For \reaA{}, the inputs are the route memory, candidate source code, proprioception and exception history.

\paragraph{Backends} implement \proc{Reset}, \proc{Render}, the policy API and a \proc{Score} read once after the last run: LIBERO/LIBERO-Pro on
MuJoCo/robosuite \citep{liu2023libero,todorov2012mujoco,mandlekar2021robosuite}, BEHAVIOR-1K on OmniGibson/Isaac
Sim \citep{li2023behavior}, RoboTwin~2.0 on SAPIEN \citep{chen2025robotwin} and a Unitree G1, where \proc{Score} is the operator's judgement.

\subsection{The Embodied Router (\syso{})}
\label{subsec:router}

The router owns every control-flow decision, with one module per coordination problem.

\begin{figure}[t]
  \centering
  \includegraphics[width=\linewidth]{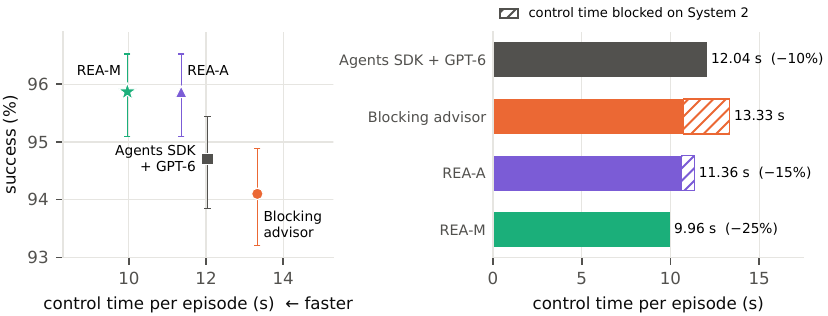}
\caption{\textbf{Task success and control time on LIBERO-Pro.} \textbf{Left:} Success rate versus control time per episode, with Wilson 95\% confidence intervals for success rates. Higher success rates and shorter control times are better. \textbf{Right:} Control time per episode. Values in brackets show the change relative to the blocking-advisor configuration and all configurations are evaluated on the full LIBERO-Pro suite with 3{,}000 episodes. Each episode additionally incurs 2.6--2.7\,s of reset time.}
  \label{fig:speed}
\end{figure}

\paragraph{Route memory: dispatch without the brain.}
Before evaluation, the router runs each candidate $p\in\mathcal{P}_t$ on $n_p$ development
episodes (seeds disjoint from test seeds, seeded-random resets) and stores the success rate
$\hat r_{t,p}$ and mean steps $\hat c_{t,p}$.  The first policy is $\arg\max_p \hat r_{t,p}$, ties broken by
lower $\hat c_{t,p}$: at test time dispatch is a lookup (the route table of \Cref{fig:system}).

\paragraph{Boundary-triggered judge gate: consult the brain when a policy returns.}
Most failures are silent: \proc{Run} returns $e=\bot$ with the task unfinished, so an exception trigger
never fires.  \reaV{} therefore calls \proc{Judge} once per policy return; \texttt{done} ends the episode and
protects completed work, \texttt{not\_done} escalates.  A free \emph{exception gate} escalates only on an
exception or truncation and never calls \syst{}.

\paragraph{Escalation ladder with a shared budget.}
The ladder $L_t=[\mathrm{first}(t),\mathrm{other}(t),\mathrm{first}(t)]$ (with three candidates, all
three in route-memory order; one rung with one) launches each escalation from the current state, and every run and brain call draws on the same
$(H-u,W-w)$, so an episode never exceeds a single-policy budget.

\paragraph{Baselines.}
A single-policy baseline is \Cref{alg:rea} with $K{=}0$; blind retry uses the unconditional gate \proc{always}.

\paragraph{Learned dispatch for VLA and code policies.}
The same \syso{} interface also supports an observation-conditioned selector. In the RoboTwin VLA study, an encoder maps the initial instruction, images and robot state to $h_0$, and a success head scores each candidate as $q_p=\sigma(w_p^\top h_0+b_p)$. The router selects $p^*=\arg\max_p q_p$ before execution and keeps that policy for the episode. Adding a code-policy head extends the seven-VLA selector to a mixed pool without changing the policy execution interface (\Cref{subsec:vla}).

\subsection{Aligned Evaluation Protocol}
\label{subsec:protocol}

Within each matched LIBERO-Pro comparison, methods use the same (task, seed) pairs, initial-state semantics, horizon and wall budget, the same frozen policy code wherever they share a library, and one final \proc{Score}. Within a campaign on one host, each episode key keeps a static GPU lane. Route memories and judge thresholds are fixed on development episodes before evaluation. The RoboTwin VLA study shares a fixed scene set and the benchmark success criterion across policy types, retaining their native action interfaces and budgets; historical baseline runs are identified separately (\Cref{subsec:vla,app:vla}).

\section{Experiments}
\label{sec:experiments}

Our experiments evaluate routing success, control efficiency, improvements across policy-generation methods, the contributions of individual modules, VLA--code integration, and transfer to other simulators and real robots.

\subsection{Setup}
\label{subsec:setup}

\paragraph{Benchmarks.}
\textbf{LIBERO-Pro}~\citep{zhou2025liberopro} perturbs LIBERO~\citep{liu2023libero} along object, position (\emph{swap}), language, task and environment axes; we use its six goal, object and spatial suites under the \emph{task} and \emph{swap} perturbations, ten tasks each, forming an out-of-distribution (OOD) test set relative to standard LIBERO.  Each test seed $s\in\{1,\dots,50\}$ uses official initial state $(s{-}1)\bmod 50$.  The \textbf{full set} is all 60 tasks $\times$ 50 seeds (3{,}000 episodes per method; the Long suite and the object, language and environment perturbations are outside our evaluation protocol). The \textbf{cross-method study} on LIBERO-Pro uses seeds 1--10 and 600 episodes.  The \textbf{matched hard subset} stress-tests recovery on six tasks (five with two distinct policies; on SS-center the \aspire{} Evo winner equals Fix) with twelve seeds each, eight of them failures of the reference Fix policy (\Cref{tab:tasks}).  On LIBERO-Pro, all methods use a 4{,}000-step horizon and a 900\,s budget; methods with routing perform at most three policy runs. BEHAVIOR-1K~\citep{li2023behavior} (R1Pro, Isaac Sim; five pick-up tasks) and RoboTwin~2.0~\citep{chen2025robotwin} (bimanual, SAPIEN) and real Unitree G1 and AlohaMini2 robots are also used to test the portability. RoboTwin has two separate evaluations: a fixed 100-scene study of VLA and code policies (\Cref{subsec:vla}) and the 200-scene transfer study (\Cref{subsec:portability}).

\paragraph{\syst{} policy-generation methods.}
\aspire{} Fix loop and Evo search~\citep{aspire2026}; Codex + GPT-6 and Agents SDK + GPT-6, coding-agent pipelines generating one program per task; and zero-shot GPT-6, two samples per task.  As an external online baseline on all three simulators we add Show-Harness~\citep{chen2026show}, a zero-shot VLM harness whose planner and controller choose every 2--4\,cm end-effector move from the camera images, run with \texttt{gpt-6-astra} (\Cref{app:showharness}).  The \emph{legacy library} holds a Codex or harness program per task and, for 59 tasks, an \aspire{} alternative; \emph{+SDK} adds the Agents SDK programs (179 programs in total).

\paragraph{Router, advisor and judge.}
Route memories use development episodes only: seeds 51--60 on the matched subset and the cross-method study, and seeds 61--70 for the full set (1{,}190 episodes for the legacy programs, 600 for the SDK ones).  \reaA{}'s dispatch agent is text only; the \emph{blocking advisor} in \Cref{subsec:speed} is a text-only agent that recommends a policy every 8 segments of 100 control steps while the router waits for it.  The visual judge is \texttt{deepseek-v4-flash-vision-exp} on the matched subset and its successor \texttt{deepseek-flash} on LIBERO-Pro, with threshold and prompt frozen on 216 development labels before any test run (\Cref{app:judge}).  For the paired LIBERO-Pro comparisons, success differences use paired exact McNemar tests \citep{mcnemar1947} with paired 95\% intervals, and time differences use task-clustered bootstrap intervals; the headline LIBERO-Pro contrasts also hold after a Holm adjustment within each question. \aspire{}'s programs are written with Codex \texttt{gpt-5.6-sol}, the other policy-generation methods use \texttt{gpt-6-astra} at high effort, \reaA{}'s router model is \texttt{jev-1.13.0}, and the caption bridge in \Cref{subsec:ablation} is \texttt{qwen3-vl-flash}.

\subsection{Routing Beats Each Policy-Generation Baseline}
\label{subsec:fullset}

\begin{table}[t]
\centering
\caption{Comparison on LIBERO-Pro full set (50 test seeds per task).  \emph{Aligned (37 tasks)}: REA uses exactly the same program files
as the Codex, \aspire{} Fix, and Agents SDK baselines.
For the remaining 23 tasks, its policy pool instead uses a
harness-generated program in place of the Codex program,
or an \aspire{} Evo program in place of the \aspire{} Fix program. The best method on development seeds 61--70 is Agents SDK (95.3\%, vs.\ 75.0\% Codex and 72.0\% \aspire{}).  Both REA routers use the exception gate (\reaM{} fails only 124 of 3{,}000 episodes).}
\label{tab:fullset}
\small
\setlength{\tabcolsep}{4pt}
\begin{tabular}{llcccc}
\toprule
 & Method & Calls/ep & All 60 (\%) & Aligned 37 (\%) & Steps/succ $\downarrow$ \\
\midrule
\multirow{4}{*}{\shortstack[l]{single\\method}} & Codex + GPT-6 & 0 & 74.4 & 81.2 & 1470 \\
  & \aspire{} Fix (59 / 36 tasks) & 0 & 65.1 & 71.9 & 1566 \\
 & Agents SDK + GPT-6 (best dev.) & 0 & 94.7 & 93.7 & 1343 \\
 & Random method per task & 0 & 78.2 & 82.5 & -- \\
\midrule
\multirow{3}{*}{\shortstack[l]{routed}} & Route memory only ($K{=}0$) & 0 & $\geq$95.7 & $\geq$95.1 & -- \\
 & \textbf{\reaM{}} (memory + ladder) & 0 & \textbf{95.9} & 95.3 & \textbf{1003} \\
 & \reaA{} (agent at dispatch) & 1.0 & \textbf{95.9} & \textbf{95.5} & 1110 \\
\midrule
ceiling & Hindsight per-task oracle & -- & 96.9 & 96.3 & -- \\
\midrule
online & Show-Harness + GPT-6 (zero-shot) & 34.4 & 2.6 & 1.8 & 31{,}875 \\
\bottomrule
\end{tabular}
\end{table}

\textbf{Routing, not a larger pool, makes the gain.}  \Cref{tab:fullset} compares \rea{} with each source method evaluated alone on the same episodes.  Choosing one method globally on development data selects Agents SDK (94.7\%); choosing a method per task at random gives 78.2\%.  Route memory chooses per task on the same development seeds and reaches 95.9\%, $+1.2$\% over the best method (86:51, $p{=}0.004$, paired 95\% CI $[+0.4,+1.9]$\%).  The comparison is exact on the 37 aligned tasks, where the pool holds only the programs used by the individual baselines: there \reaM{} reaches 95.3\% against 93.7\% (54:25, $p{=}0.001$, CI $[+0.6,+2.5]$\%), within 1.0\% of the hindsight oracle over the whole pool; on the other 23 tasks it is on par with the best method (96.8 vs.\ 96.3\%).  It also costs less: 1{,}003 steps per success versus 1{,}343 ($-25\%$), because each task goes to its most reliable policy on development episodes (the SDK program on only 14 of the 60 tasks), and those policies are shorter.  \textbf{Per program, routing uses less data than policy generation.}  Route memory uses 10 development seeds per program (1{,}790 episodes) and no model tokens, whereas each generation pipeline spent 1{,}249--1{,}328 development episodes and 29--89M input tokens on its own library (\Cref{app:pool}).  \emph{Route memory is as good as an agent at dispatch}: \reaA{} reaches the same 95.9\% (26:26 paired) and, as \Cref{subsec:speed} shows, is slower. \textbf{Putting the brain in the control loop does not substitute for policies.}  Show-Harness, which asks \texttt{gpt-6-astra} for every move from the images, solves 78 of the 3{,}000 episodes (2.6\%); 2{,}806 of them end at the 900\,s budget, after 34 model calls on average.

\begin{figure}[t]
  \centering
  \includegraphics[width=\linewidth]{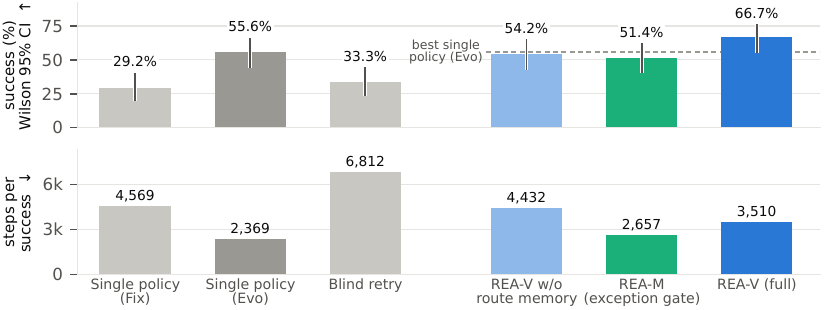}
  \caption{\textbf{Router ablation} on the matched hard subset (6 tasks $\times$ 12 seeds $=$ 72 episodes per method; identical initial states, GPU lanes and policy code).  Without route memory the judge ladder is Fix, Evo, Fix; blind retry re-runs Fix once.}
  \label{fig:ablation}
\end{figure}

\subsection{Route Memory Reduces Waiting for the Agent}
\label{subsec:speed}

\Cref{fig:speed} measures the designs of \Cref{fig:timeline} on one host and one policy library (numbers in \Cref{tab:speed}). \textbf{Waiting for the brain is expensive.}  An agent advisor that the router waits for, although consulted only 1.7 times per episode, leaves the control path idle 2.62\,s per episode (20\% of its control time), delays the first action by 1.5\,s, and is slower than one SDK program (13.33 vs.\ 12.04\,s) with no gain in success.  Its advice also costs success, nearly all at the first dispatch: it overrides the policy selected by route memory in 1{,}236 episodes, on which that policy alone solves 1{,}168 and the advisor 1{,}124 (rest: 1{,}708 vs.\ 1{,}699). \textbf{Route memory removes the wait at its source.}  \reaM{} has the lowest control time among the timed methods and ties \reaA{} for the highest success: $-3.37$\,s ($-25\%$, CI $[-31,-20]\%$) with $+1.8$\% of success against the blocking advisor (90:37, $p{<}10^{-5}$), and $-2.08$\,s ($-17\%$, CI $[-23,-11]\%$) against Agents SDK + GPT-6 in its own campaign.  The first gap is not hardware: both methods execute a control step in 10.3--10.4\,ms outside \syst{} waits. \reaA{}, whose agent reads the same route memory, matches its success but is 1.39\,s slower per paired episode: 0.73\,s waiting for its selection call, the rest because it starts a different program in 46\% of episodes.

\begin{figure}[t]
  \centering
  \includegraphics[width=\linewidth]{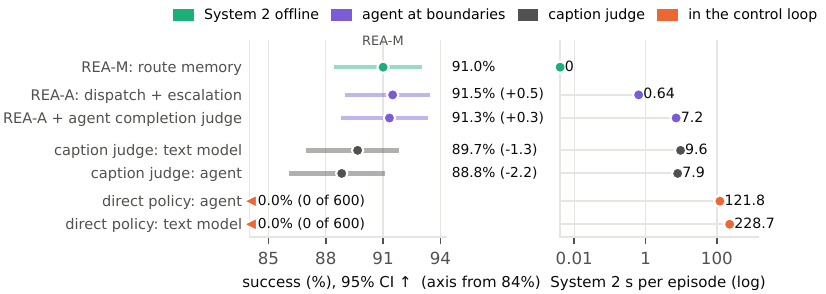}
  \caption{\textbf{Where should \syst{} sit?}  LIBERO-Pro seeds 1--10 (600 episodes per method).  Left: success with Wilson 95\% intervals; the two direct policies solve no episode (arrows at the axis edge).  Right: seconds per episode spent in \syst{} calls (agent, caption model or text judge), log scale.  The direct policies act through 3\,cm / 15$^\circ$ Cartesian IK steps and a gripper command, at most 80 decisions, from scene captions.  Calls and paired counts in \Cref{tab:placement}.}
  \label{fig:placement}
\end{figure}

\subsection{\rea{} Improves Policies from Different Methods}
\label{subsec:amplify}

On \aspire{}'s Fix-loop policies (LIBERO-Pro seeds 1--10, 590 episodes; \Cref{tab:amplify}), \reaM{} alone lifts success from 66.3\% to 71.9\% ($+5.6$\%, 49:16, $p{<}10^{-4}$) while \emph{reducing} steps per success (1{,}467 vs.\ 1{,}557): it escalated on an exception only 6 times although 166 episodes failed, so nearly all of the gain is route memory running the Evo policy first where development favoured it.  The judge gate (\reaV{}) adds a further $+3.1$\% (37:19, $p{=}0.022$), for 74.9\% in total, at 2.1$\times$ the wall time per success (\Cref{fig:pareto}). This paired contrast is the gate's net effect; how it splits into rescued failures, undone successes and replay variance is only approximate, because \reaM{}'s own run stands in for \reaV{}'s first run (\Cref{tab:mechanism}).  On zero-shot GPT-6 policies, a threefold weaker library, \reaV{} lifts success from 19.3\% to 26.3\% (50:8, $p{<}10^{-4}$); the comparison is unchanged on the 59 tasks shared with \aspire{} (50:8), so the task excluded by the sandbox does not drive it.

\subsection{Where Do the Gains Come From?}
\label{subsec:ablation}

The matched subset concentrates on episodes where single policies fail (\Cref{fig:ablation}).  The full router solves 48/72 episodes and is the only method to outperform the best single policy ($+11.1$\%, $p{=}0.039$ unadjusted); the gate contrast below is the one that survives a Holm correction (adjusted $p{=}0.014$, against 0.12 for this gain). \textbf{Both modules contribute.}  Without route memory the judge must find the better policy by trial and error (39/72, $p{=}0.078$).  Replacing the judge gate by the exception gate costs 15.3\% (12:1, $p{=}0.003$): of the twelve episodes only the full router solves, seven are failed first runs rescued by escalation and five are replay variance.  \textbf{The comparison is at a matched escalation budget}: \reaM{}'s cap $K{=}1$ never binds, because all three of its escalations ran to the 4{,}000-step horizon, so with $K{=}2$ it would behave identically on every episode.  The subset is selected for Fix failures; on unselected seeds the same gate contrast is $+3.1$\% (\Cref{subsec:amplify}). \textbf{Uncoordinated recovery does not help}: blind retry barely beats its own first policy (24 vs.\ 21, $p{=}0.66$).

\paragraph{Where should \syst{} sit?}  \Cref{fig:placement} moves the agent through every online role on one library and host.  Asking it to pick the policies at dispatch and recovery (\reaA{}) changes success by $+0.5$\% (6:3, $p{=}0.51$) while adding a selection call.  A completion judge that reads scene captions, whether the agent or a text model, costs 1.3--2.2\% and spends 7.9--9.6\,s per episode in \syst{} calls, since on this strong library the exception gate suffices; as a direct Cartesian policy it solves none of 600 episodes, a second failing interface after the one-token probe (\Cref{app:onetoken}).  The brain pays off offline and, where policies fail silently, as a visual judge.

\paragraph{Why can a noisy judge help?}  Its final verdicts are barely more accurate than always answering \texttt{done} (0.72 vs.\ 0.67 matched) or less (0.67 vs.\ 0.75 LIBERO-Pro), but their \texttt{done} decisions are precise (0.83, 0.905), and re-running a completed task usually keeps it completed: by the proxy attribution of \Cref{tab:mechanism}, re-runs convert 24\% (19\%) of failed first runs and undo 5\% (9\%) of successful ones.  Measured on final verdicts only, this is a consistent explanation, not a tested mechanism; the $+15.3$\% of this subset is a stress test, and $+3.1$\% on unselected seeds (\Cref{subsec:amplify}) is the estimate to expect.

\subsection{VLA and Code Policies Share System 0}
\label{subsec:vla}

\textbf{Learned policies run in the same embodied runtime.} We evaluate \rea{} on 50 RoboTwin~2.0 tasks, with one clean and one randomized scene per task (100 episodes). \rea{} (VLA policies) selects among seven VLA models; \rea{} (VLA policies + Code policies) adds a Python code policy to the same candidate pool. Both use a learned \syso{} selector that predicts success from the initial instruction, camera images and robot state, chooses the highest-scoring policy, and retains it for the episode. VLA policies execute native action chunks; the code policy generates and executes a Cartesian gripper command at each decision. This directly instantiates both policy types as \sysz{} under the common runtime interface (\Cref{app:vla}).

\begin{table}[!t]
\centering
\caption{\textbf{Comparison with VLA and code policies on RoboTwin~2.0.} Each row evaluates the same 50 clean and 50 randomized scenes. Success counts and overall rates are shown; best per column is bold. Historical baselines and the VLA extension were run in different windows; MEDex uses a separate host. Protocols and policy budgets are detailed in \Cref{app:vla}.}
\label{tab:robotwin_vla}
\small
\setlength{\tabcolsep}{5pt}
\begin{tabular}{lccc}
\toprule
Method & Clean & Randomized & Overall (\%) \\
\midrule
\multicolumn{4}{l}{\emph{Agent and code baselines (historical runs)}} \\
Codex CLI & 28/50 & 27/50 & 55 \\
OpenAI Agents SDK & 22/50 & 24/50 & 46 \\
\aspire{} (RoboTwin port) & 25/50 & 26/50 & 51 \\
Show-Harness & 5/50 & 1/50 & 6 \\
\rea{} (Code policies) & 27/50 & 28/50 & 55 \\
\midrule
\multicolumn{4}{l}{\emph{VLA and mixed policies}} \\
MEDex (single VLA) & 42/50 & 38/50 & 80 \\
\rea{} (VLA policies) & \textbf{44/50} & 40/50 & 84 \\
\rea{} (VLA policies + Code policies) & \textbf{44/50} & \textbf{42/50} & \textbf{86} \\
\bottomrule
\end{tabular}
\end{table}

\begin{samepage}
\textbf{Combining VLA and code policies improves success.} \Cref{tab:robotwin_vla} shows 84\% success with VLA policies and 86\% with the mixed pool, compared with 80\% for the MEDex single-VLA baseline. Both routed VLA configurations solve 44/50 clean scenes. Adding code raises randomized successes from 40/50 to 42/50, with two additional successes and no lost VLA-only successes. These results provide direct evidence for VLA execution and episode-level dispatch across learned and programmatic policies under the same \rea{} interface. Policy choices and training details are in \Cref{app:vla}.
\end{samepage}

\subsection{Extensional Experiments on BEHAVIOR and RoboTwin-2}
\label{subsec:portability}

\begin{figure}[t]
  \centering
  \includegraphics[width=\linewidth]{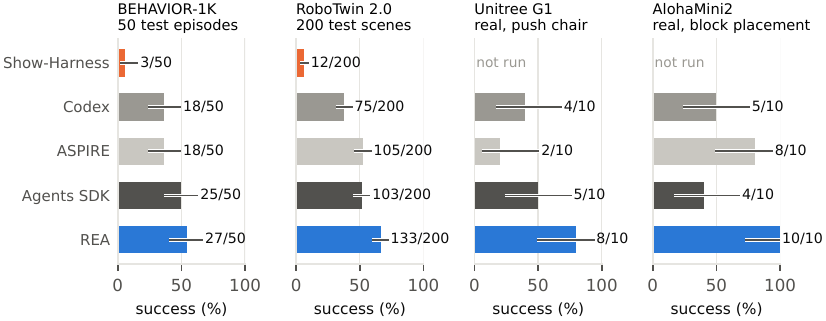}
  \caption{\textbf{Results of portability experiments.} Bars show success rates; whiskers indicate Wilson 95\% confidence intervals. BEHAVIOR-1K: test instances 1--10 of five R1Pro pick-up tasks (50 episodes), 1{,}500\,s per episode, \texttt{gpt-6-astra} for every method; Codex + GPT-6 and \aspire{} are frozen programs (the \aspire{}-style Fix loop), Agents SDK and Show-Harness are online agents. RoboTwin~2.0: the two test sets (200 scenes, 50 tasks, half clean and half randomized), at most 40 decisions per episode. Unitree G1 of real walk-and-push-chair trials and AlohaMini2 of real block-placement trials are both manually reviewed. Show-Harness was not run on either real robot due to the low success rates. Details in \Cref{app:behavior,app:g1chair,app:aloha-agents}.}
  \label{fig:portability}
\end{figure}

On BEHAVIOR-1K (R1Pro, Isaac Sim), which needed only a new backend, \rea{} solves 27 of 50 test episodes over five pick-up tasks, the most of the compared methods: 25 for the online Agents SDK agent with the same model, tools and budget, 18 each for the Codex and \aspire{}-style Fix programs, and 3 for Show-Harness (\Cref{fig:portability}, \Cref{tab:behavior}).  Against the Agents SDK agent the margin is not significant (7:5, $p{=}0.77$) and comes from one task, \texttt{camera} (5 against 1); against the frozen-program baselines it is 13:4 ($p{=}0.049$) and 12:3 ($p{=}0.035$).

In the separate 200-scene RoboTwin~2.0 transfer study, a fast reflex model serves as the \sysz{} policy and the router hands single decisions to a deliberate model.  Run side by side with the Codex CLI on the 200 test scenes, \rea{} solves 97 against 75 (42:20, $p{=}0.007$).  Run again on a later night it solves 133, against 105 for \aspire{}, 103 for the Agents SDK agent and 12 for Show-Harness ($p{<}0.001$ each); the same configuration moved by 36 scenes between the two nights, so only the side-by-side pair compares methods without a window effect (\Cref{app:behavior}).

\subsection{Real Robot Experiments on Unitree-G1}
\label{subsec:unitree-g1}

On a Unitree G1 humanoid that must walk to a chair and push it, with the same model and \sysz{} policies for every method and success judged on site, \rea{} succeeds in 8 of 10 trials, against 5 for the Agents SDK agent, 4 for Codex and 2 for \aspire{} ($p{=}0.023$; \Cref{tab:g1chair}). Figure~\ref{fig:showcase} shows \rea{} episodes on all three simulators and on the real robot, including a LIBERO-Pro test episode in which the judge gate climbs the full escalation ladder.

\subsection{Real-World Block Placement on AlohaMini2}
\label{subsec:aloha}

\begin{figure}[t]
  \centering
  \begingroup%
\setlength{\unitlength}{1pt}%
\definecolor{alohaBaseline}{RGB}{119,151,176}%
\colorlet{alohaGrid}{black!14}%
\resizebox{\linewidth}{!}{%
\begin{picture}(530,180)
\sffamily\fontsize{11}{13}\selectfont
\put(0,172){\makebox(0,0)[l]{\bfseries (a) Agent methods}}
\put(284,172){\makebox(0,0)[l]{\bfseries (b) VLA + Code}}

\put(80,46){\color{alohaGrid}\rule{0.4pt}{106pt}}
\put(146,46){\color{alohaGrid}\rule{0.4pt}{106pt}}
\put(212,46){\color{alohaGrid}\rule{0.4pt}{106pt}}
\put(360,46){\color{alohaGrid}\rule{0.4pt}{106pt}}
\put(426,46){\color{alohaGrid}\rule{0.4pt}{106pt}}
\put(492,46){\color{alohaGrid}\rule{0.4pt}{106pt}}

\put(0,144.5){\makebox(0,0)[l]{Codex}}
\put(80,138){\color{alohaBaseline}\rule{66pt}{13pt}}
\put(222,144.5){\makebox(0,0)[l]{5/10}}
\put(0,117.5){\makebox(0,0)[l]{Agents SDK}}
\put(80,111){\color{alohaBaseline}\rule{52.8pt}{13pt}}
\put(222,117.5){\makebox(0,0)[l]{4/10}}
\put(0,90.5){\makebox(0,0)[l]{ASPIRE}}
\put(80,84){\color{alohaBaseline}\rule{105.6pt}{13pt}}
\put(222,90.5){\makebox(0,0)[l]{8/10}}
\put(0,63.5){\makebox(0,0)[l]{\bfseries REA}}
\put(80,57){\color{infrecred}\rule{132pt}{13pt}}
\put(222,63.5){\makebox(0,0)[l]{\bfseries 10/10}}

\put(284,130.5){\makebox(0,0)[l]{$\pi_{0.5}$}}
\put(360,124){\color{alohaBaseline}\rule{39.6pt}{13pt}}
\put(502,130.5){\makebox(0,0)[l]{3/10}}
\put(284,86){\makebox(0,0)[l]{\bfseries REA}}
\put(284,72){\makebox(0,0)[l]{\fontsize{10}{11}\selectfont ($\pi_{0.5}$ + Code)}}
\put(360,76){\color{infrecred}\rule{105.6pt}{13pt}}
\put(502,82.5){\makebox(0,0)[l]{\bfseries 8/10}}

\put(80,42){\color{black!55}\rule{132pt}{0.4pt}}
\put(360,42){\color{black!55}\rule{132pt}{0.4pt}}
\put(80,39){\color{black!55}\rule{0.4pt}{3pt}}
\put(146,39){\color{black!55}\rule{0.4pt}{3pt}}
\put(212,39){\color{black!55}\rule{0.4pt}{3pt}}
\put(360,39){\color{black!55}\rule{0.4pt}{3pt}}
\put(426,39){\color{black!55}\rule{0.4pt}{3pt}}
\put(492,39){\color{black!55}\rule{0.4pt}{3pt}}
\put(80,28){\makebox(0,0){0}}
\put(146,28){\makebox(0,0){50}}
\put(212,28){\makebox(0,0){100}}
\put(360,28){\makebox(0,0){0}}
\put(426,28){\makebox(0,0){50}}
\put(492,28){\makebox(0,0){100}}
\put(146,10){\makebox(0,0){Success rate (\%)}}
\put(426,10){\makebox(0,0){Success rate (\%)}}
\end{picture}%
}%
\endgroup
  \caption{\textbf{Experimetal results of AlohaMini2 block placement.} Success rates in two separate studies, with ten trials per method; labels give successful trials. (a) Four agent methods. (b) $\pi_{0.5}$ alone and \rea{} routing between $\pi_{0.5}$ and Code. The two studies retain their own protocols and timing measures (\Cref{app:aloha-agents,app:aloha-controllers}).}
  \label{fig:aloha}
\end{figure}
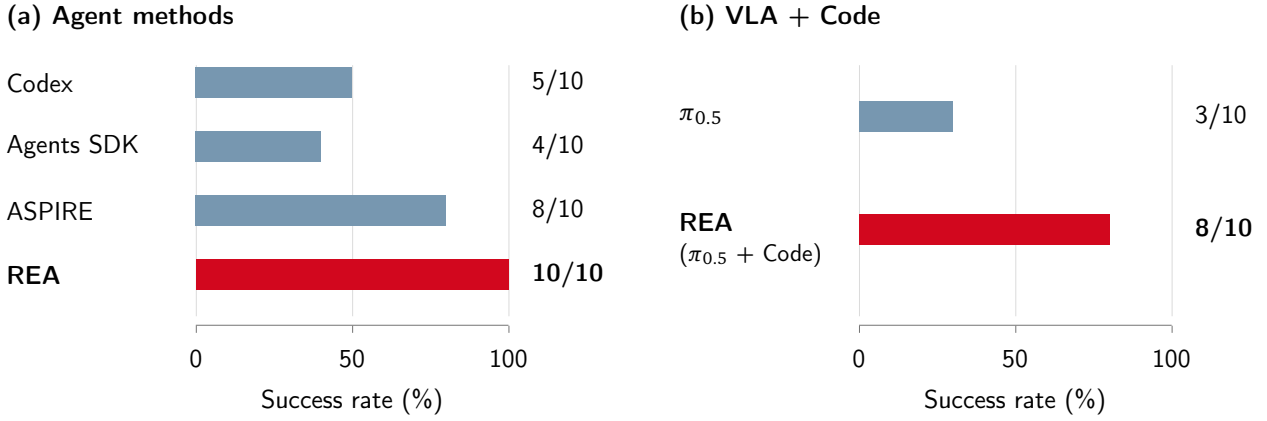

We evaluate the AlohaMini2 right arm on grasping an object outside a box, releasing it inside, and leaving it stable after the gripper retracts. In a 40-trial comparison with ten valid trials per method, \rea{} completes 10/10 trials, compared with 8/10 for \aspire{}, 5/10 for Codex and 4/10 for Agents SDK (\Cref{fig:aloha}). Each trial allows up to three grasp attempts without a fixed total time limit, and outcomes use the original manual review (\Cref{app:aloha-agents}).

In a separate 20-trial controller study, \rea{} routes asynchronously between $\pi_{0.5}$ and a Code controller and succeeds in 8/10 trials, versus 3/10 for $\pi_{0.5}$ alone. Placement failures fall from five to zero, while both groups have two grasp failures; Code executes in five of the ten \rea{} trials. The groups were collected on different days with their original session settings (\Cref{app:aloha-controllers}). \rea{} has higher observed success and longer mean recorded time in both studies: 1{,}618.0\,s versus 1{,}314.9\,s for \aspire{} in the four-method study, and 128.89\,s versus 16.57\,s of active time for $\pi_{0.5}$ alone in the controller study (\Cref{tab:aloha_agents,tab:aloha_vla}).

\section{Discussion}
\label{sec:discussion}

\paragraph{Use the brain ahead of time or at policy returns, never inside a running policy.}
Dispatching from route memory outperforms each individual policy-generation method and is as accurate as asking an agent at every dispatch, at no online cost (\Cref{tab:fullset}, \Cref{fig:placement}); consulted in the loop, the brain adds waiting (\Cref{fig:speed}); consulted only when a policy returns, a noisy judge adds success where policies fail silently, at extra time (\Cref{tab:amplify}, \Cref{fig:ablation}).  The brain still sets the ceiling (success rates differ by 20--47\% across methods), and the router decides how much of it is reached.  The gate should be chosen per library on development episodes: the free exception gate suffices when policies rarely fail, the judge gate pays for its calls when they fail silently.

\paragraph{One interface, different selectors.}
The shared \proc{Run} interface supports both route-memory lookup and learned policy selection. In the RoboTwin~2.0 extension, a learned selector chooses one of seven VLAs or a Python policy from the initial observation and keeps that choice fixed throughout the episode (\Cref{subsec:vla}, \Cref{app:vla}). This study evaluates the shared policy interface and initial selection; recovery-gate effects are evaluated separately on LIBERO-Pro.


\section{Conclusion}
\label{sec:conclusion}

\rea{} separates an embodied system into a \syst{} brain, an embodied router and fast \sysz{} policies under one aligned runtime.  On 60 LIBERO-Pro tasks it beats each method that contributes policies to its pool and is the fastest timed pipeline without any online brain call.  Its gains come from the router rather than a larger pool: route memory dispatches as well as an online agent at no model cost, and the boundary-triggered judge helps where policies fail silently.  The same interfaces carry over to BEHAVIOR-1K, RoboTwin~2.0 and a real Unitree G1, where \rea{} solves the most episodes.  The separate 100-episode RoboTwin~2.0 study demonstrates VLA and code policies within the same runtime: \rea{} reaches 84\% with VLA policies and 86\% with VLA+code, compared with 80\% for MEDex, the strongest single VLA baseline.  We hope \rea{} becomes a shared platform for developing and comparing embodied brains and fast policies.

\clearpage
\phantomsection
\addcontentsline{toc}{section}{References}
\bibliography{references}

\newpage
\appendix
\raggedbottom
\crefalias{section}{appendix}
\crefalias{subsection}{appendix}
\section{LIBERO-Pro: Additional Results}
\label{app:grp-libero}

\subsection{Speed Details}
\label{app:speed}

\begin{table}[tp]
\centering
\caption{Per-suite success (of 500) and mean control time (s) on the full set.  \reaM{} is faster than the blocking advisor in every suite and at least as accurate in every suite.}
\label{tab:speed-suite}
\small
\begin{tabular}{lcccc}
\toprule
 & \multicolumn{2}{c}{Blocking advisor} & \multicolumn{2}{c}{\reaM{}} \\
\cmidrule(lr){2-3}\cmidrule(lr){4-5}
Suite & Succ & Control & Succ & Control \\
\midrule
goal\_swap & 456 & 14.30 & 470 & 11.32 \\
goal\_task & 435 & 14.19 & 442 & 9.94 \\
object\_swap & 474 & 10.06 & 499 & 7.68 \\
object\_task & 481 & 12.36 & 486 & 7.43 \\
spatial\_swap & 491 & 14.75 & 491 & 11.02 \\
spatial\_task & 486 & 14.34 & 488 & 12.38 \\
\bottomrule
\end{tabular}
\end{table}

\emph{Control time} runs from the end of the environment reset to the end of the episode, includes every policy run and every \syst{} call the control path waits for, and is charged against the 900\,s budget; \emph{wait} is the part spent blocked on \syst{}.  All timed methods in \Cref{tab:speed} ran on one host and execute a control step at the same rate outside \syst{} waits (9.5--10.4\,ms), so \reaM{}'s 3.37\,s advantage over the blocking advisor is its 2.62\,s wait plus 0.75\,s from 83 fewer control steps per episode.  Intervals resample the 60 tasks (4{,}000 bootstrap draws).  In 12 blocking-advisor episodes no advice completed and the policy selected by route memory ran alone; they are kept.  \Cref{tab:placement} gives the calls and paired counts behind \Cref{fig:placement}.

\begin{table}[tp]
\centering
\caption{Speed on the LIBERO-Pro full set, ordered by where \syst{} runs (3{,}000 episodes per method on one host; the routed methods share one 179-program library).  Control: control time per episode, excluding reset; $\Delta$: paired change in control time and $b$:$c$: episodes solved only by the method in the row : only by the blocking advisor; Wait: time per episode the control path is blocked on \syst{}.}
\label{tab:speed}
\small
\setlength{\tabcolsep}{3.5pt}
\begin{tabular}{llcccccc}
\toprule
Method & \syst{} runs & Calls/ep & Succ.\ (\%) $\uparrow$ & $b$:$c$ & Control (s) $\downarrow$ & $\Delta$ & Wait (s) $\downarrow$ \\
\midrule
Agents SDK + GPT-6 & offline & 0 & 94.7 & 109:91 & 12.04 & $-10\%$ & 0 \\
\midrule
Blocking advisor & in the loop & 1.71 & 94.1 & -- & 13.33 & -- & 2.62 \\
\reaA{} & at dispatch & 1.00 & \textbf{95.9} & 97:44 & 11.36 & $-15\%$ & 0.73 \\
\textbf{\reaM{}} & offline & 0 & \textbf{95.9} & 90:37 & \textbf{9.96} & $\mathbf{-25\%}$ & 0 \\
\bottomrule
\end{tabular}
\end{table}

\begin{table}[tp]
\centering
\caption{Where should \syst{} sit (\Cref{fig:placement})?  LIBERO-Pro seeds 1--10 (600 episodes per method, one host, legacy library, exception gate unless stated).  $b$:$c$ against \reaM{}; S2 time: seconds per episode spent in \syst{} calls (agent, caption model or text judge).  The per-step policies act through 3\,cm / 15$^\circ$ Cartesian IK steps and a gripper command, at most 80 decisions, from scene captions.}
\label{tab:placement}
\small
\setlength{\tabcolsep}{3.5pt}
\begin{tabular}{llcccc}
\toprule
\syst{} placement & Configuration & Calls/ep & S2 time (s) & Success (\%) & $b$:$c$ \\
\midrule
offline & \reaM{} (route memory) & 0 & 0 & 91.0 & -- \\
dispatch + recovery & \reaA{} & 0.99 & 0.64 & \textbf{91.5} & 6:3 \\
+ completion & \reaA{} + agent judge & 4.18 & 7.18 & 91.3 & 12:10 \\
\midrule
caption judge & text model & 3.13 & 9.57 & 89.7 & 11:19 \\
 & agent & 3.23 & 7.94 & 88.8 & 9:22 \\
\midrule
direct policy & agent & 52.9 & 121.8 & 0.0 & 0:546 \\
 & text model & 90.1 & 228.7 & 0.0 & 0:546 \\
\bottomrule
\end{tabular}
\end{table}

\subsection{Same-Pool Baselines and Development Cost}
\label{app:pool}

\Cref{tab:fullset} compares \rea{} with each source method evaluated alone on the same 3{,}000 episodes.  The \emph{random method per task} row is the expected success of choosing one of the three methods uniformly per task; the \emph{hindsight oracle} picks, for each task, the method with the most successes on test episodes, a ceiling no selector fitted on development data reaches in expectation.  \emph{Route memory only} counts the 16 escalated episodes of \reaM{} as failures: their first run raised an exception, and 5 of them succeeded only after escalation, so $K{=}0$ reaches at least 2{,}871 of 3{,}000.  \emph{Aligned tasks}: on 37 tasks the pool holds exactly the programs the single-method rows ran, byte-identical for Codex and Agents SDK and the same program files for \aspire{} Fix (one task has no \aspire{} program in either); there the oracle is over the whole pool.  On the other 23 tasks the pool's Codex slot holds a harness-generated program (14 tasks) or its \aspire{} slot an \aspire{} Evo winner (10 tasks; one task has both), programs no single-method row ran.  \Cref{tab:devcost} compares the development budgets.

\begin{table}[tp]
\centering
\caption{Development budgets on LIBERO-Pro (60 tasks).  Program-generation episodes are the development evaluations the coding agents ran while generating their programs; route-memory episodes replay frozen programs on seeds 61--70.}
\label{tab:devcost}
\small
\setlength{\tabcolsep}{4pt}
\begin{tabular}{lccc}
\toprule
Component & Dev.\ episodes & Input tokens & Dev.\ success of result \\
\midrule
Codex + GPT-6 program generation & 1{,}328 & 89.0M & 75.0\% (600 ep.) \\
Agents SDK + GPT-6 program generation & 1{,}249 & 28.8M & 95.3\% (600 ep.) \\
\aspire{} Fix loop and Evo search & not recorded & not recorded & 72.0\% (590 ep.) \\
\midrule
\rea{} route memory (all 179 programs) & 1{,}790 & 0 & -- \\
\bottomrule
\end{tabular}
\end{table}

\subsection{LIBERO-Pro Cross-Method Study}
\label{app:pro10}

\Cref{tab:amplify} gives the two libraries evaluated in \Cref{subsec:amplify} with their latency.  On zero-shot policies \reaV{} routes over both samples; on \aspire{} policies over the Fix programs and, on the 11 tasks that have one, the Evo winner.  Under the exception gate an escalation is rare (0.01 per episode); under the judge gate it happens about once per episode, which is where \reaV{}'s extra wall time comes from.  One \aspire{} task is excluded because its program calls \texttt{getattr}, which the sandbox refuses.

\begin{table}[tp]
\centering
\caption{LIBERO-Pro seeds 1--10: \rea{} on policies generated by two \syst{} methods (\Cref{subsec:amplify}).  $b$:$c$ = episodes solved only by \rea{} : only by the better single library above it (paired exact McNemar, all $p{<}10^{-4}$).  Wall time includes judge latency; J = judge, Esc.\ = escalations.  \aspire{} rows cover 590 episodes.  Best per group in bold.}
\label{tab:amplify}
\footnotesize
\setlength{\tabcolsep}{3pt}
\begin{tabular}{lccccccc}
\toprule
Method & Succ & Rate (\%) & Steps/succ & Wall/succ (s) & J calls/ep & Esc./ep & $b$:$c$ \\
\midrule
GPT-6 zero-shot, sample 1 & 116/600 & 19.3 & \textbf{4299} & \textbf{127} & 0 & 0 & -- \\
GPT-6 zero-shot, sample 2 & 91/600 & 15.2 & 5315 & 481 & 0 & 0 & -- \\
\quad + \reaV{} (both samples) & \textbf{158/600} & \textbf{26.3} & 6023 & 317 & 2.45 & 1.45 & 50:8 \\
\midrule
\aspire{} Fix-loop policies & 391/590 & 66.3 & 1557 & \textbf{40.9} & 0 & 0 & -- \\
\aspire{} Evo (11 tasks) & 57/110 & 51.8 & 2529 & 78.4 & 0 & 0 & -- \\
\quad + \reaM{} (exception gate) & 424/590 & 71.9 & \textbf{1467} & 57.9 & 0 & 0.01 & 49:16 \\
\quad + \reaV{} (judge gate) & \textbf{442/590} & \textbf{74.9} & 2513 & 120.7 & 2.10 & 1.10 & 73:22 \\
\bottomrule
\end{tabular}
\end{table}

\begin{figure}[tp]
  \centering
  \includegraphics[width=\linewidth]{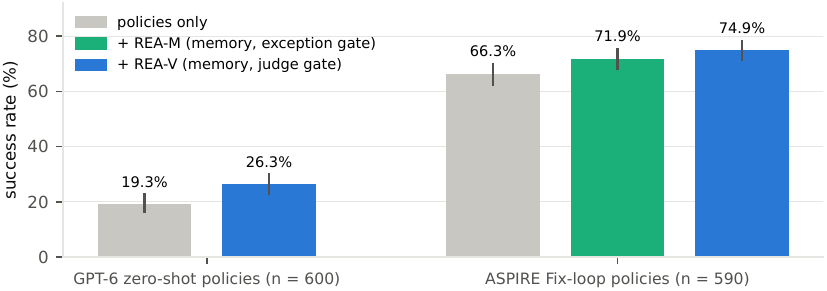}
  \caption{\rea{} improves policy libraries of very different quality (\Cref{tab:amplify}; bars are Wilson 95\% intervals).}
  \label{fig:amplify}
\end{figure}

\subsection{Success, Cost and Latency}
\label{app:pareto}

\begin{figure}[tp]
  \centering
  \includegraphics[width=\linewidth]{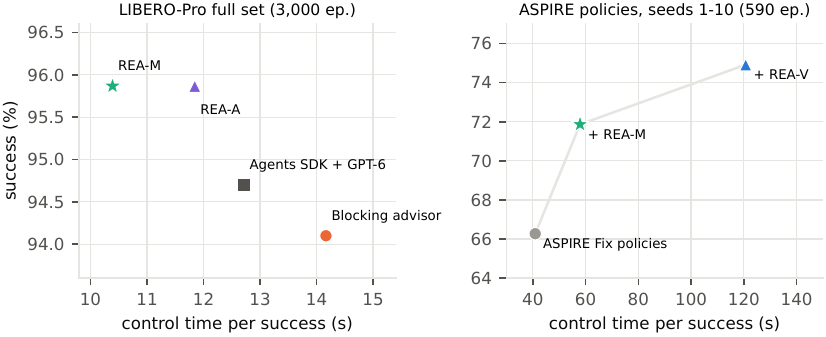}
  \caption{Success against control time per success.  Left: LIBERO-Pro full set (one host; up and to the left is better).  Right: \aspire{} policies on seeds 1--10 (\Cref{tab:amplify}).}
  \label{fig:pareto}
\end{figure}

\Cref{fig:pareto} places each method by success and control time per success.  On the full set \reaM{} ties for the highest success and is the cheapest timed method.  On \aspire{} policies each router module moves the single policy along a trade-off: route memory adds success at lower cost per success, and the judge gate adds more success at about twice the time per success.

\section{The Router: Episode Loop, Matched Subset and Judge}
\label{app:grp-router}

\subsection{Episode Loop}
\label{app:alg}

\Cref{alg:rea} gives the episode loop of the judge- and exception-gated configurations (\Cref{subsec:router}).  Route memory fixes the ladder before the reset.  Each run receives only the remaining step and wall budget, so escalations never extend the episode, and the judge's latency is charged to the same budget.  The success predicate is read once, by \proc{Score}, after the agent stops.

\begin{algorithm}[tp]
\SetAlgoLined
\DontPrintSemicolon
\SetKwComment{tcp}{$\triangleright$\ }{}
\KwIn{task $t$, seed $s$, route memory $\mathcal{M}$, judge $J$, gate $g\in\{\proc{judge},\proc{exception},
  \proc{always}\}$, escalations $K$, budgets $H,W$}
$L \leftarrow \mathcal{M}.\mathrm{ladder}(t)$;\ \ $\proc{Reset}(s)$;\ \ $f_0 \leftarrow \proc{Render}()$;\ \
  $u,w \leftarrow 0$ \tcp*{\syso{} lookup}
\For{$k \leftarrow 0$ \KwTo $K$}{
  $(\Delta,\delta,e,\tau) \leftarrow \proc{Run}(L[\min(k,|L|-1)];\ H-u,\ W-w)$;\ \ $u \mathrel{+}= \Delta$;\ \
    $w \mathrel{+}= \delta$ \tcp*{\sysz{}}
  \lIf{$g=\proc{judge}$}{$(v,\delta_J) \leftarrow J(f_0,\proc{Render}(),\ell_t)$;\ \ $w \mathrel{+}= \delta_J$
    \tcp*[f]{\syst{}}}
  \lIf{$k=K$ \textbf{or} $u\ge H$ \textbf{or} $w\ge W$}{\textbf{break}}
  \lIf{($g=\proc{judge}$ \textbf{and} $v=\texttt{done}$) \textbf{or}
    ($g=\proc{exception}$ \textbf{and} $e=\bot$ \textbf{and not} $\tau$)}{\textbf{break}}
}
\Return $\proc{Score}()$ \tcp*{read once, evaluation only}
\caption{One \rea{} episode with the judge or exception gate.  Nothing before the last line reads the success predicate, rewards or object poses; the verdict after the last run is only logged.}
\label{alg:rea}
\end{algorithm}

\subsection{Matched Hard Subset: Tasks, Methods and Totals}
\label{app:setup}

The matched subset in \Cref{subsec:ablation} takes six LIBERO-Pro tasks on which \aspire{}'s Fix policy often fails, with twelve test seeds each, eight of them Fix failures (\Cref{tab:tasks}).  Every method in the router ablation runs the loop of \Cref{alg:rea} on the same initial states, GPU lanes and policy code, and differs only in the first policy, the ladder and the gate (\Cref{tab:method-configs}); \Cref{tab:ablation} gives the totals.

\begin{table}[tp]
\centering
\caption{Tasks of the matched hard subset.  Five tasks have two distinct candidate policies (\aspire{} Fix and Evo); on SS-center the Evo winner is byte-identical to Fix, so its two columns are two replays of the same program and differ only by replay variance.}
\label{tab:tasks}
\setlength{\tabcolsep}{3pt}
\small
\begin{tabular}{llccc}
\toprule
Suite & Task (abbreviated) & Seeds & Fix/12 & Evo/12 \\
\midrule
libero\_goal\_swap & Open drawer, put bowl (GS-drawer) & 31--39, 42, 44, 47 & 3 & 11 \\
libero\_goal\_task & Open drawer, put cream cheese (GT-drawer) & 31--42 & 4 & 5 \\
libero\_object\_task & Pick up butter, place in basket (OT-butter) & 31--39, 46, 47, 50 & 2 & 2 \\
libero\_spatial\_swap & Bowl between plate and ramekin (SS-between) & 8, 14, 31--40 & 4 & 12 \\
libero\_spatial\_swap & Bowl from table center (SS-center) & 31--41, 43 & 3 & 7 \\
libero\_spatial\_swap & Bowl on ramekin (SS-ramekin) & 31--40, 43, 44 & 5 & 3 \\
\bottomrule
\end{tabular}
\end{table}

\begin{table}[tp]
\centering
\caption{Method configurations of the router ablation (\Cref{tab:ablation}).  All methods run the episode loop of \Cref{alg:rea}; blind retry uses the unconditional gate \proc{always}. }
\label{tab:method-configs}
\small
\begin{tabular}{lllcc}
\toprule
Method & First policy & Ladder & Gate & Max escalations $K$ \\
\midrule
Single policy (Fix / Evo) & Fix / Evo & [Fix] / [Evo] & -- & 0 \\
Blind retry & Fix & [Fix, Fix] & always & 1 \\
\reaV{} w/o route memory & Fix & [Fix, Evo, Fix] & judge & 2 \\
\reaM{} & route memory & route memory & exception & 1 \\
\textbf{\reaV{} (full)} & \textbf{route memory} & \textbf{route memory} & \textbf{judge} & \textbf{2} \\
\bottomrule
\end{tabular}
\end{table}

\begin{table}[tp]
\centering
\caption{Router ablation on the matched hard subset (\Cref{fig:ablation}; 6 tasks $\times$ 12 seeds $=$ 72 episodes per method; Wilson intervals; identical initial states, GPU lanes and policy code).}
\label{tab:ablation}
\small
\setlength{\tabcolsep}{3pt}
\begin{tabular}{lcccccc}
\toprule
Method & Memory & Gate & Succ/72 & Succ.\ (\%) $\uparrow$ & 95\% CI & Steps/succ $\downarrow$ \\
\midrule
Single policy (Fix) & -- & -- & 21 & 29.2 & [19.9, 40.5] & 4569 \\
Single policy (Evo) & -- & -- & 40 & 55.6 & [44.1, 66.5] & 2369 \\
Blind retry & -- & always & 24 & 33.3 & [23.5, 44.8] & 6812 \\
\reaV{} w/o route memory & -- & judge & 39 & 54.2 & [42.7, 65.2] & 4432 \\
\reaM{} & \checkmark & exception & 37 & 51.4 & [40.1, 62.6] & 2657 \\
\textbf{\reaV{} (full)} & \checkmark & judge & \textbf{48} & \textbf{66.7} & [55.2, 76.5] & 3510 \\
\bottomrule
\end{tabular}
\end{table}

\subsection{Per-Task Results on the Matched Hard Subset}
\label{app:pertask}

\begin{figure}[tp]
  \centering
  \includegraphics[width=\linewidth]{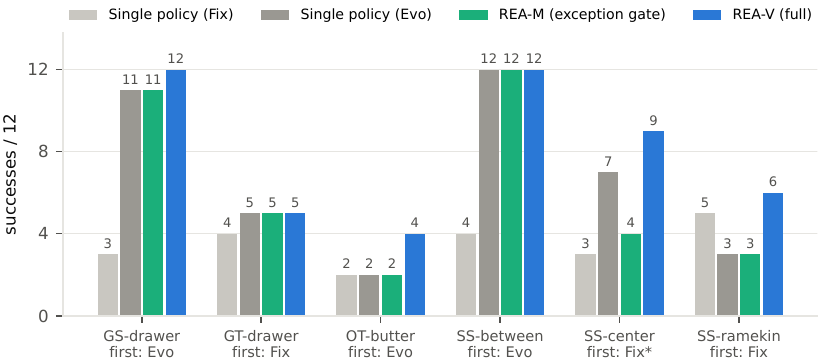}
  \caption{Successes out of 12 per task; the policy that route memory dispatches first is shown under each task (* SS-center has a single distinct policy).  \reaV{} is best or tied-best on every task.}
  \label{fig:pertask}
\end{figure}

The full router is best or tied-best on every task (\Cref{fig:pertask,tab:app-pertask}).  Where route memory dispatches the stronger policy first (GS-drawer, SS-between) the router matches or exceeds that policy; on the other tasks its gains come from escalating after a failed first run, at the cost of more steps per episode.

\begin{table}[tp]
\centering
\caption{Full per-task statistics (steps include the 10 settle steps; escalation steps are the steps of every run after the first).  On SS-center, Fix and Evo are the same program.}
\label{tab:app-pertask}
\setlength{\tabcolsep}{2.5pt}
\footnotesize
\begin{tabular}{llccccc}
\toprule
Task & Method & Succ/12 & Steps/ep & Steps/succ & Judge calls/ep & Escalation steps/ep \\
\midrule
\multirow{6}{*}{GS-drawer} & Single policy (Fix) & 3 & 2312 & 9246 & 0 & 0 \\
 & Single policy (Evo) & 11 & 2138 & 2332 & 0 & 0 \\
 & \reaV{} w/o route memory & 7 & 3581 & 6138 & 1.92 & 1267 \\
 & Blind retry & 6 & 3987 & 7974 & 0 & 1684 \\
 & \textbf{\reaV{} (full)} & 12 & 3846 & 3846 & 1.92 & 1682 \\
 & \reaM{} (exception gate) & 11 & 2138 & 2332 & 0 & 0 \\
\midrule
\multirow{6}{*}{GT-drawer} & Single policy (Fix) & 4 & 2579 & 7738 & 0 & 0 \\
 & Single policy (Evo) & 5 & 2630 & 6313 & 0 & 0 \\
 & \reaV{} w/o route memory & 5 & 4000 & 9600 & 2.08 & 1310 \\
 & Blind retry & 5 & 3958 & 9499 & 0 & 1310 \\
 & \textbf{\reaV{} (full)} & 5 & 4000 & 9600 & 2.08 & 1310 \\
 & \reaM{} (exception gate) & 5 & 2865 & 6876 & 0 & 282 \\
\midrule
\multirow{6}{*}{OT-butter} & Single policy (Fix) & 2 & 1200 & 7197 & 0 & 0 \\
 & Single policy (Evo) & 2 & 1213 & 7279 & 0 & 0 \\
 & \reaV{} w/o route memory & 3 & 3230 & 12{,}920 & 2.92 & 2037 \\
 & Blind retry & 3 & 2175 & 8700 & 0 & 967 \\
 & \textbf{\reaV{} (full)} & 4 & 3092 & 9276 & 2.92 & 1879 \\
 & \reaM{} (exception gate) & 2 & 1213 & 7279 & 0 & 0 \\
\midrule
\multirow{6}{*}{SS-between} & Single policy (Fix) & 4 & 591 & 1774 & 0 & 0 \\
 & Single policy (Evo) & 12 & 647 & 647 & 0 & 0 \\
 & \reaV{} w/o route memory & 11 & 1066 & 1163 & 1.83 & 462 \\
 & Blind retry & 1 & 1136 & 13{,}628 & 0 & 512 \\
 & \textbf{\reaV{} (full)} & 12 & 647 & 647 & 1.00 & 0 \\
 & \reaM{} (exception gate) & 12 & 647 & 647 & 0 & 0 \\
\midrule
\multirow{6}{*}{SS-center} & Single policy (Fix) & 3 & 632 & 2529 & 0 & 0 \\
 & Single policy (Evo) & 7 & 599 & 1026 & 0 & 0 \\
 & \reaV{} w/o route memory & 9 & 1066 & 1422 & 1.92 & 497 \\
 & Blind retry & 6 & 1178 & 2356 & 0 & 556 \\
 & \textbf{\reaV{} (full)} & 9 & 1076 & 1435 & 1.75 & 444 \\
 & \reaM{} (exception gate) & 4 & 637 & 1911 & 0 & 0 \\
\midrule
\multirow{6}{*}{SS-ramekin} & Single policy (Fix) & 5 & 682 & 1638 & 0 & 0 \\
 & Single policy (Evo) & 3 & 669 & 2677 & 0 & 0 \\
 & \reaV{} w/o route memory & 4 & 1460 & 4379 & 2.25 & 785 \\
 & Blind retry & 3 & 1191 & 4763 & 0 & 545 \\
 & \textbf{\reaV{} (full)} & 6 & 1380 & 2760 & 2.17 & 697 \\
 & \reaM{} (exception gate) & 3 & 692 & 2769 & 0 & 0 \\
\bottomrule
\end{tabular}
\end{table}

\subsection{Route Memory}
\label{app:route}

Route memory is fitted on development seeds 51--60 (seeded-random resets), never on test seeds.  For each task it ranks the candidate policies by development success, breaks ties by fewer steps per success, and stores the first policy and the ladder [first, other, first] (\Cref{tab:route-full}).

\begin{table}[tp]
\centering
\caption{Route memory of the matched subset.  Evo is dispatched first exactly where development shows it to be clearly stronger.  On SS-center the Evo program is byte-identical to Fix, so both development configurations count as Fix evidence (6/20) and the ladder collapses to [Fix, Fix].}
\label{tab:route-full}
\small
\setlength{\tabcolsep}{3.5pt}
\begin{tabular}{llccll}
\toprule
Task & Suite & Fix dev & Evo dev & First & Ladder \\
\midrule
GS-drawer & goal\_swap & 1/10 & 10/10 & Evo (higher rate) & Evo, Fix, Evo \\
GT-drawer & goal\_task & 1/10 & 1/10 & Fix (tie; fewer steps) & Fix, Evo, Fix \\
OT-butter & object\_task & 0/10 & 3/10 & Evo (higher rate) & Evo, Fix, Evo \\
SS-between & spatial\_swap & 1/10 & 10/10 & Evo (higher rate) & Evo, Fix, Evo \\
SS-center & spatial\_swap & 6/20 & = Fix & Fix (single policy) & Fix, Fix \\
SS-ramekin & spatial\_swap & 4/10 & 2/10 & Fix (higher rate) & Fix, Evo, Fix \\
\bottomrule
\end{tabular}
\end{table}

\subsection{Visual-Judge Calibration}
\label{app:judge}

\begin{figure}[tp]
  \centering
  \includegraphics[width=\linewidth]{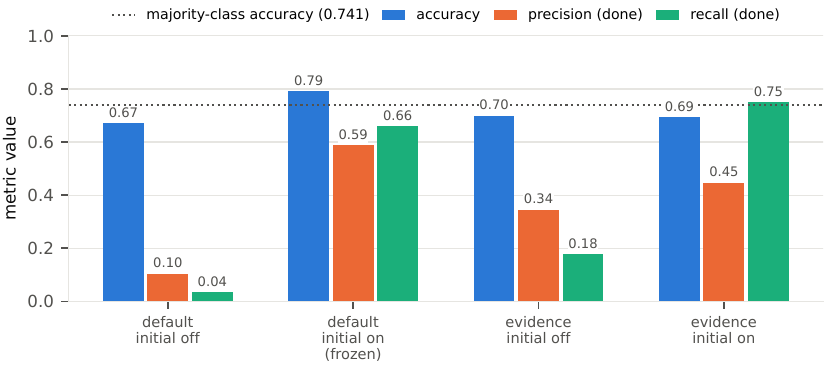}
  \caption{Judge calibration on development labels at threshold 0.6 (\texttt{done} is the positive class).  With the default prompt, showing the episode-initial frame lifts the judge above the majority-class baseline; without it, precision and recall collapse.}
  \label{fig:judge-cal}
\end{figure}

The visual judge is calibrated on development data only: 216 labelled episodes (seeds 51--62, 56 successes and 160 failures) whose initial and final 800$\times$512 agentview frames were saved, with labels from the final score.  We sweep the prompt style and whether the initial frame is shown, at thresholds 0.5--0.9, and select by accuracy, then precision (\Cref{fig:judge-cal,tab:judge-sweep}). The successor model used on LIBERO-Pro scores 0.782 on the same 216 labels (0.792 for the frozen snapshot).

\begin{table}[tp]
\centering
\caption{Development calibration sweep (\texttt{deepseek-v4-flash-vision-exp}, threshold 0.6; the result is flat over 0.5--0.9).  The frozen configuration is in bold.}
\label{tab:judge-sweep}
\small
\begin{tabular}{llccccc}
\toprule
Prompt & Initial frame & Accuracy & Precision & Recall & Latency (s) & Cost/call (USD) \\
\midrule
default & off & 0.671 & 0.105 & 0.036 & 1.60 & 0.000247 \\
\textbf{default} & \textbf{on} & \textbf{0.792} & \textbf{0.587} & 0.661 & 1.60 & 0.000390 \\
evidence & off & 0.699 & 0.345 & 0.179 & 1.93 & 0.000353 \\
evidence & on & 0.694 & 0.447 & 0.750 & 2.01 & 0.000527 \\
\bottomrule
\end{tabular}
\end{table}

\subsection{Mechanism of the Judge Gate}
\label{app:mechanism}

\Cref{tab:mechanism} attributes the episodes of the full router, using \reaM{} (same first policy, seed and lane) as a proxy for the outcome of the first run.  On test episodes, confusion matrices of the final verdict against the final score are TP 35 / FP 7 / FN 13 / TN 17 on the matched subset (done-precision 0.83) and TP 276 / FP 29 / FN 166 / TN 119 on LIBERO-Pro (done-precision 0.905).  These count only the verdict after the last run, not every verdict that triggered an escalation.

\begin{table}[tp]
\centering
\caption{Episode attribution of the full router.  \reaM{} stands in for the first-run outcome, so the split of the paired wins into rescues and replay variance ($12 = 7 + 5$; $37 = 28 + 9$) is approximate.}
\label{tab:mechanism}
\small
\setlength{\tabcolsep}{4pt}
\begin{tabular}{p{7.7cm}cc}
\toprule
Attribution (proxy $=$ \reaM{} outcome) & Matched (72) & LIBERO-Pro (590) \\
\midrule
Rescued: first run failed $\rightarrow$ escalated $\rightarrow$ success & 7 & 28 \\
Not rescued: first run failed $\rightarrow$ escalated $\rightarrow$ still failed & 22 & 117 \\
Survived: first run succeeded $\rightarrow$ judged not done $\rightarrow$ re-run $\rightarrow$ success & 20 & 198 \\
Harmed: first run succeeded $\rightarrow$ re-run $\rightarrow$ failed & 1 & 19 \\
Forfeited: first run failed $\rightarrow$ judged done $\rightarrow$ stopped & 1 & 12 \\
Kept: first run succeeded and judged done & 16 & 207 \\
Replay variance: proxy failed but the router's own first run succeeded & 5 & 9 \\
\midrule
Conversion of failures, rescued / (rescued + not rescued) & 24\% & 19\% \\
Destruction of successes, harmed / (harmed + survived) & 5\% & 9\% \\
Paired result vs.\ \reaM{} & $+12/{-}1$, $p{=}0.003$ & $+37/{-}19$, $p{=}0.022$ \\
\bottomrule
\end{tabular}
\end{table}

\section{Other Simulators and Online Baselines}
\label{app:grp-sim}

\subsection{RoboTwin VLA and Code Policy Integration}
\label{app:vla}

\paragraph{Scenes and execution.}
The VLA study uses a fixed set of 100 RoboTwin~2.0 scenes: 50 tasks, each with one clean and one randomized scene, taken from episode 0 of the scene manifest. It preserves each scene's seed, instruction and initial configuration. These scenes differ from the 200-scene transfer study, which uses episodes 3 and 4. They were used in earlier REA development and serve here as a fixed-scene reevaluation; they are excluded from fitting and selecting the new code head. Both policy types run in the same live simulator and use the official task-success criterion. VLA policies execute their native action chunks and task-specific action limits. The code policy receives four RGB views, robot state, calibration and execution feedback; GPT-6 Astra generates Python that computes one Cartesian \texttt{move\_gripper} target per decision, with at most 40 decisions per episode.

\paragraph{Selector and training.}
A Qwen3.5-4B encoder with LoRA reads the initial instruction, three camera views and robot state. Seven sigmoid heads predict the success of $\pi_{0.5}$, GigaBrain, Spatial Forcing, X-VLA, XWAM, PatchWAM and MEDex. The mixed configuration adds a linear code-policy head, trained with binary cross-entropy on execution success/failure labels while the encoder, LoRA and seven VLA heads remain frozen. The reported checkpoint uses 247 training scenes and 40 development scenes; epoch 4 is selected by mixed-policy success on the development scenes, with code-head loss breaking ties. Later completed calibration scenes are not used to refit this checkpoint. The selector picks the highest predicted success probability at the initial observation and caches that choice for the entire episode; each selected policy continues to react to new observations internally.

\paragraph{Comparison scope.}
\Cref{tab:robotwin_vla} combines the new VLA configurations with earlier baselines on the same 100 scenes. Codex CLI, OpenAI Agents SDK, the RoboTwin port of \aspire{}, and code-only \rea{} are historical runs. The code-only \rea{} row is the earlier rule-based reflex/deliberate router, not the new Python candidate alone. MEDex is a separate-host rerun with 42 clean and 38 randomized successes. Show-Harness retains its own 100-step budget; the marked historical result uses an adapter later found to accumulate orientation updates incorrectly. The corrected adapter has only a partial run and is not represented by that row. The table reports success rates across these runs; timing comparisons are confined to matched campaigns.

\paragraph{Contribution of the code candidate.}
The mixed selector chooses code on nine randomized scenes, succeeding on eight. On the other 91 scenes, its selected VLA and outcome match the VLA-only run. Relative to VLA-only routing, code adds successes on \texttt{blocks\_ranking\_size} and \texttt{put\_object\_cabinet}, with no lost successes. Thus the measured benefit comes from choosing complementary policy types before execution; this study does not enable mid-episode switching or a visual judge gate.

\subsection{BEHAVIOR-1K and RoboTwin 2.0}
\label{app:behavior}

On BEHAVIOR-1K the three-level contract, the episode runner and the final score (\aspire{}'s own \texttt{task\_completed()}, read once after the agent stops) are unchanged from LIBERO; only the backend differs.  Tasks: \texttt{trash} (pick up the soda can), \texttt{radio} (the radio), \texttt{teddy} (the teddy bear), \texttt{drill} (the power drill on a workbench) and \texttt{camera} (the digital camera on the floor), R1Pro, 2025-challenge instances, Isaac Sim 4.5 and OmniGibson 3.7.2; \aspire{}'s own release has configurations only for \texttt{trash} and \texttt{radio}.

\paragraph{Test instances 1--10 (\Cref{tab:behavior}).}  One RTX PRO 5000 host, a 1{,}500\,s program budget per episode enforced at the API boundary (model latency, tool execution and image rendering all count), the same 16 R1Pro tools and the same domain text for every online agent.  \rea{} runs the routed policy fixed on RoboTwin before any BEHAVIOR episode (one policy per decision, at most 40 decisions, a \texttt{finish(arm)} accepted only after an in-hand check); the Agents SDK agent runs one loop of at most 60 turns over the same tools.  Both use \texttt{gpt-6-astra} at medium reasoning and ran interleaved on the same simulator slots, under a protocol written before either ran.  Codex + GPT-6 is a frozen program written once by the Codex CLI (\texttt{gpt-6-astra}, high effort; the better of two development samples) and \aspire{} the same program after up to two rounds of \aspire{}'s trace-guided Fix loop on development instances 26--35, run by the Codex CLI (\texttt{gpt-6-astra}, medium); both ran two weeks earlier on the same host and budget, so those pairs are descriptive. Show-Harness is in \Cref{app:showharness}.

\paragraph{RoboTwin 2.0 transfer study (200 scenes).}  Two test sets of the frozen 500-scene manifest (episodes 3 and 4 of every task; 100 scenes each, 50 clean and 50 randomized) in SAPIEN, with \texttt{gpt-6-astra} at medium reasoning, at most 2{,}048 output tokens per request and at most 40 action decisions per episode for every method.  With no policy library, a fast reflex model is \rea{}'s \sysz{} policy and the router hands single decisions to a deliberate model; its configuration was fixed on development scenes before any test episode.  \aspire{} is the harness owner's RoboTwin port.  Paired against \rea{}'s later run (133), \rea{} alone : the compared method alone is 46:16 against the Agents SDK agent, 40:12 against \aspire{}, 66:8 against Codex and 121:0 against Show-Harness ($p{<}0.001$ each); the side-by-side run against Codex is 42:20 ($p{=}0.007$).

\begin{table}[tp]
\centering
\caption{BEHAVIOR-1K, test instances 1--10 (successes of 10 per task; mean program time per episode).  Every method uses \texttt{gpt-6-astra}.  Paired against \rea{} (\rea{} alone : the compared method alone): Agents SDK 7:5 ($p{=}0.77$), Codex 13:4 ($p{=}0.049$), \aspire{}-style Fix 12:3 ($p{=}0.035$), Show-Harness 25:1 ($p{<}10^{-5}$).  Best per column in bold.}
\label{tab:behavior}
\small
\setlength{\tabcolsep}{3.2pt}
\begin{tabular}{lccccccc}
\toprule
Method & trash & radio & teddy & drill & camera & Total & Prog.\ s/ep \\
\midrule
Codex + GPT-6 (frozen program) & 6 & 6 & 5 & 1 & 0 & 18/50 & 994 \\
\aspire{}-style Fix (frozen program) & 7 & 6 & 4 & 1 & 0 & 18/50 & \textbf{853} \\
Show-Harness (online, scripted approach) & 0 & 2 & 1 & 0 & 0 & 3/50 & 1492 \\
OpenAI Agents SDK agent (online) & \textbf{9} & \textbf{8} & \textbf{7} & 0 & 1 & 25/50 & 1049 \\
\textbf{\rea{}} (routed episode runtime) & \textbf{9} & 7 & 5 & \textbf{1} & \textbf{5} & \textbf{27/50} & 1122 \\
\bottomrule
\end{tabular}
\end{table}

\paragraph{Development instances 26--35 of \texttt{trash} and \texttt{radio} (\Cref{tab:behavior-dev}).}  An earlier two-task setting (one H20 GPU per lane, 2{,}000\,s per policy run, 3{,}600\,s per episode) in which \rea{} replays the frozen policy that route memory ranks first among three written by \aspire{}'s interactive agent ($K{=}0$).  These are development-stage latency numbers; the agent's program-generation time is not counted.

\begin{table}[tp]
\centering
\caption{BEHAVIOR-1K development instances 26--35.  Elapsed time includes reset and scene loading (for the interactive agent, per completed trial).  The frozen policy is the best of three candidates on these instances, so its success is optimistic.  Best per column in bold.}
\label{tab:behavior-dev}
\small
\setlength{\tabcolsep}{3.5pt}
\begin{tabular}{llcccc}
\toprule
Method (development instances 26--35) & Task & Succ & Steps/succ & Wall/succ (s) & Elapsed/succ (s) \\
\midrule
Human-written reference program & trash & 5/10 & 7795 & \textbf{1445} & \textbf{1565} \\
Interactive agent (agent in the loop) & trash & 4/10 & -- & -- & 6438 \\
\rea{}, frozen policy ($K{=}0$) & trash & 7/10 & \textbf{6887} & 1611 & 1734 \\
\midrule
Human-written reference program & radio & 2/10 & 27{,}947 & 5547 & 5848 \\
Interactive agent (agent in the loop) & radio & \textbf{4/10} & -- & -- & 7281 \\
\rea{}, frozen policy ($K{=}0$) & radio & \textbf{4/10} & \textbf{9193} & \textbf{1945} & \textbf{2097} \\
\bottomrule
\end{tabular}
\end{table}

\subsection{Show-Harness Baseline}
\label{app:showharness}

Show-Harness~\citep{chen2026show} (commit \texttt{5a3713b}) is a zero-shot VLM harness: a subgoal planner splits the task into per-arm stages, and a controller prompt asks the model, at every step, for one discrete move per arm (\texttt{MV\_FWD}, \texttt{MV\_DOWN}, \ldots, \texttt{GRASP}, \texttt{RELEASE}, \texttt{DONE}) from a front view and the wrist views, with its plugins for proprioception, move memory, grasp recovery and variable step size.  We run its code unchanged with \texttt{gpt-6-astra} and supply only the embodiment adapter its authors leave to each lab (views, token directions, begin pose, gripper widths); development episodes set the adapter and never enter the results. \textbf{LIBERO-Pro}: single-arm zero-shot mode on the full set (3{,}000 episodes, the protocol of \Cref{tab:fullset}), high effort, at most 200 controller decisions, 2\,cm fine and 4\,cm coarse moves; 78 successes (2.6\%), 2{,}806 episodes end at the 900\,s budget (episodes that ended on a provider failure were rerun).  \textbf{RoboTwin~2.0}: dual-arm zero-shot mode, medium effort, its own 100-step budget; 12 of 200. \textbf{BEHAVIOR-1K}: Show-Harness assumes a fixed-base arm at a table, while the R1Pro starts metres from the object, so the approach is scripted before Show-Harness starts (the same search and navigation policies as the other methods, no model call, inside the budget), its \texttt{GRASP} runs the harness's planned grasp, and moves are inverse-kinematics steps.

\subsection{An Agent as Its Own Cerebellum: the One-Token VLM Baseline}
\label{app:onetoken}

To test whether \syst{} can also serve as \sysz{}, we measured a direct vision-language policy on LIBERO: the same VLM used as our visual judge (\texttt{deepseek-v4-flash-vision-exp}) receives a 256-pixel RGB montage (external and wrist camera), the task instruction and proprioception, and emits exactly one action token per decision from a 15-token Cartesian/gripper vocabulary; each token is applied for five 20\,Hz control steps (\Cref{tab:nano}).

\begin{table}[tp]
\centering
\caption{One-token VLM baseline (official test initial states; this policy uses its own 600-step horizon.}
\label{tab:nano}
\small
\begin{tabular}{lrrl}
\toprule
Condition & Episodes completed & Successes & Rate \\
\midrule
Pro (6 suites, 600-step horizon) & 1{,}987 & 1 & 0.0005 \\
\bottomrule
\end{tabular}
\end{table}

\paragraph{Diagnosis.}  Two lateral tokens account for 89\% of all outputs and the descend token is almost never chosen, so the end effector never comes within 25\,cm of the target; the single success is a lateral sweep that flips a stove knob.  The cause is grounding, not actuation: a scripted controller with privileged object poses but the same tokens, step size and horizon completes the pick-and-place tasks on 7 of 7 seeds, while the model cannot map world axes onto either camera view.  A richer interface (3\,cm / 15$^\circ$ Cartesian IK steps, a gripper command, scene captions, up to 80 decisions; \Cref{fig:placement}) also solves 0 of 600 LIBERO-Pro episodes.  The VLM is therefore used where it is reliable, a before/after judgement at a policy boundary, and continuous control is left to \sysz{} policies.

\section{Real-Robot Experiments}
\label{app:grp-g1}

\subsection{Unitree G1: Walk to and Push a Chair}
\label{app:g1chair}

A Unitree G1 humanoid, secured by a gantry, starts in front of an office chair and receives the instruction \emph{take two small steps forward, push the chair, then retract the arms and stand still}.  Every method uses the same robot, the same model (\texttt{deepseek-flash}, low reasoning) and the same four \sysz{} policies: walk forward (0.25\,m/s for 1\,s), the robot's native two-arm forward push, arm release and stop.  \aspire{} writes one program with a single model request (code as policy); Codex + \texttt{deepseek-flash}, run as a separate campaign, writes a program with the Codex CLI; the Agents SDK agent calls the policies as tools (2--3 model requests); \rea{} dispatches each policy at a policy boundary from the head and wrist camera images (4 decisions).  Chair contact and displacement are not measured automatically, so \proc{Score} is the operator's judgement on site, recorded after each trial.

\begin{table}[tp]
\centering
\caption{Unitree G1 walk-and-push-chair task, operator-judged success.  Fisher exact tests against \rea{}; \rea{} shown in \Cref{fig:g1chair}.}
\label{tab:g1chair}
\small
\begin{tabular}{lccc}
\toprule
Method & Successful trials & Success (\%) & $p$ vs.\ \rea{} \\
\midrule
\aspire{} (code as policy) & 2/10 & 20 & 0.023 \\
Codex + \texttt{deepseek-flash} (Codex CLI) & 4/10 & 40 & 0.17 \\
OpenAI Agents SDK (tool agent) & 5/10 & 50 & 0.35 \\
\textbf{\rea{}} & \textbf{8/10} & \textbf{80} & -- \\
\bottomrule
\end{tabular}
\end{table}

\begin{figure}[p]
  \centering
  \includegraphics[width=\linewidth,height=0.84\textheight,keepaspectratio]{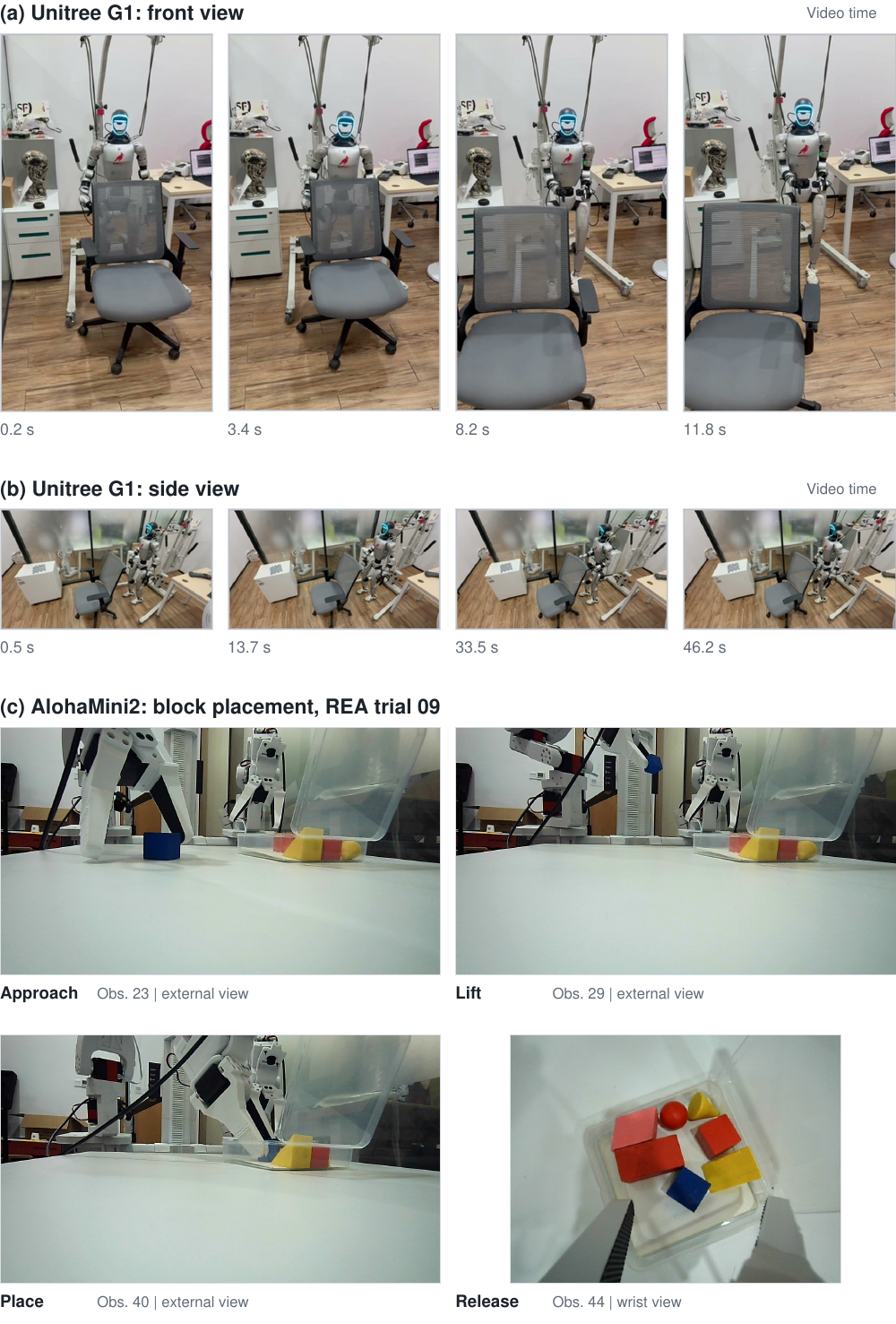}
  \caption{\textbf{\rea{} on real robots.} (a,b) Unitree G1 walks up to a chair and pushes it, shown from front and side external cameras; labels give time in each video. (c) AlohaMini2 block placement in a successful trial from the four-method study (trial 09): approach, lift, place and release, read left to right and then top to bottom. The first three images are external views; the last is the wrist view. Aloha labels give recorded observation numbers.}
  \label{fig:g1chair}
\end{figure}

\subsection{Unitree G1: Latency Check}
\label{app:g1}

Our Unitree G1 service, which does not use the router, computes brain decisions ahead of the request: a one-token vision decision (\texttt{deepseek-v4-flash-vision-exp}) as the brain, an in-process fused safety gate (robot state, RealSense depth and Livox lidar vetoes) and locomotion/arm actions as the executor.  In \emph{pipeline} mode a decision loop runs ahead of requests and the service returns the newest decision, 0.04--0.79\,s old, which the gate re-checks against a live depth capture at execution (15--40\,ms, not part of the request latency).  \Cref{tab:g1} reports medians over 5--8 trials on the robot's onboard computer (Jetson) with live sensors.  This measures latency only, not task success.

\begin{table}[tp]
\centering
\caption{Request latency on a Unitree G1 (medians over 5--8 trials).  The two right columns are the same service; only pipelining differs.  The pipeline total is request latency; the motion RPC (7--45\,ms) and the live depth re-read (15--40\,ms) follow at execution.}
\label{tab:g1}
\small
\setlength{\tabcolsep}{4pt}
\begin{tabular}{lcccc}
\toprule
Stage & Original bridge & Fast harness & Service, single-shot & Service, pipeline \\
\midrule
Capture & 1.1--1.7\,s & 18\,ms & 14--36\,ms & off the request path \\
Model & 1.8--2.4\,s & 1.1--1.5\,s & 0.80--1.25\,s & off the request path \\
Fused gate & 2.3\,s & 4--10\,ms & 0.6--2.2\,ms & 0.6--2.2\,ms \\
Motion accepted & 1.5\,s & 7--45\,ms & 7--45\,ms & 7--45\,ms (at execution) \\
\midrule
Request latency & $\approx$5\,s & 1.17--1.9\,s & 0.95\,s & \textbf{1.0--3.6\,ms} \\
\bottomrule
\end{tabular}
\end{table}

\subsection{AlohaMini2: Four-Method Comparison}
\label{app:aloha-agents}

We evaluate the AlohaMini2 right arm on moving an object from outside a box into it. Success requires releasing the object and retracting the gripper while the object remains stable inside. Codex, Agents SDK, \aspire{} and \rea{} each have ten valid, manually reviewed trials. Matching trial numbers use paired layouts, with the objects and robot reset before each trial. Each trial allows at most three grasp attempts and has no total time limit. The first \aspire{} trial uses \texttt{deepseek-flash}; the other 39 trials use \texttt{openai/gpt-6-astra}. One successful trial is shown in \Cref{fig:g1chair}(c).

\begin{table}[tp]
\centering
\caption{AlohaMini2 four-method study (40 valid trials). Elapsed time is the recorded \texttt{elapsed\_s}. All means cover ten trials, including failures, except the successful-trial time column.}
\label{tab:aloha_agents}
\small
\setlength{\tabcolsep}{3.5pt}
\begin{tabular}{lrrrrrr}
\toprule
Method & Success & Rate (\%) & \shortstack{Elapsed (s)\\all trials} & \shortstack{Elapsed (s)\\successful} & \shortstack{Actions\\per trial} & \shortstack{Model calls\\per trial} \\
\midrule
Codex & 5/10 & 50 & 1118.2 & 1699.2 & 13.6 & 32.1 \\
Agents SDK & 4/10 & 40 & 1079.5 & 1205.4 & 24.5 & 38.1 \\
\aspire{} & 8/10 & 80 & 1314.9 & 1222.2 & 30.1 & 35.7 \\
\textbf{\rea{}} & \textbf{10/10} & \textbf{100} & 1618.0 & 1618.0 & 34.1 & 41.1 \\
\bottomrule
\end{tabular}
\end{table}

\begin{table}[H]
\centering
\caption{AlohaMini2 VLA and Code study. All time statistics cover ten trials. Active time is \texttt{activeMs}/1{,}000; elapsed time is \texttt{endedAt}-\texttt{startedAt} and includes pauses.}
\label{tab:aloha_vla}
\small
\setlength{\tabcolsep}{3.5pt}
\begin{tabular}{lrrrrrr}
\toprule
Method & Success & Rate (\%) & \shortstack{Active (s)\\mean} & \shortstack{Active (s)\\median} & \shortstack{Elapsed (s)\\mean} & \shortstack{Failures\\grasp / place} \\
\midrule
$\pi_{0.5}$ & 3/10 & 30 & 16.57 & 16.34 & 29.36 & 2 / 5 \\
\rea{} ($\pi_{0.5}$ + Code) & \textbf{8/10} & \textbf{80} & 128.89 & 46.29 & 132.70 & 2 / 0 \\
\bottomrule
\end{tabular}
\end{table}

\rea{} completes all ten trials, compared with eight for \aspire{}, five for Codex and four for Agents SDK. Its higher observed success accompanies longer mean runtime and more actions and model calls (\Cref{tab:aloha_agents}); early termination of failed trials affects the runtime averages.

\subsection{AlohaMini2: VLA and Code Policies}
\label{app:aloha-controllers}

A separate 20-trial study compares $\pi_{0.5}$ alone with \rea{} routing between the same step-5{,}000 checkpoint and a Code controller. Both execute action blocks of 15 targets at 30\,Hz; $\pi_{0.5}$ predicts 50 steps per inference. All ten \rea{} trials use asynchronous routing. The $\pi_{0.5}$ trials were collected on October 26, 2026 under the earlier D1 setup; the \rea{} trials were collected on October 27 under \texttt{cube-move-on-approach-v1}. Task prompts and pauses retain their original session settings. Two invalid $\pi_{0.5}$ records are excluded, leaving ten trials in each group.

The $\pi_{0.5}$ group dispatches 272 action blocks. \rea{} dispatches 449 $\pi_{0.5}$ blocks and 116 Code blocks; Code executes in five of the ten trials, while all ten count as the routed configuration. \rea{} completes eight trials versus three for $\pi_{0.5}$ alone. Placement failures fall from five to zero, with two grasp failures in each group. Both mean active time and mean elapsed time are longer for \rea{} (\Cref{tab:aloha_vla}).

\end{document}